\documentclass[10pt]{article}

\usepackage{authblk}
\usepackage{natbib}
\usepackage{mathtools}
\usepackage{color}
\usepackage{amsmath}
\usepackage{amsthm}
\usepackage{amssymb}
\usepackage{amsbsy}
\usepackage{amsfonts}
\usepackage{amscd}
\usepackage{mathrsfs}
\usepackage{bm}
\usepackage{breqn}
\usepackage{color, soul}
\usepackage{enumerate}
\usepackage{empheq}
\usepackage{rotating}
\usepackage{ftnxtra}
\usepackage{fnpos}
\usepackage[titletoc,toc,title]{appendix}
\usepackage{euscript}
\usepackage{graphicx}
\usepackage{epsfig}
\usepackage{epstopdf}
\DeclareGraphicsExtensions{.pdf,.png,.jpg,.eps}
\usepackage{pstool}
\usepackage{upgreek}
\usepackage{mathrsfs}
\usepackage{subfig}
\usepackage{verbatim}
\usepackage{fancyhdr} 

\usepackage{psfrag}

\numberwithin{equation}{section}
\usepackage[ruled,linesnumbered]{algorithm2e}  
\usepackage{amsmath, amssymb}
\theoremstyle{plain}	
\newtheorem{thm}{Theorem}[section]

\newtheorem*{prop*}{Proposition}
\theoremstyle{definition}	

\newtheorem{remark}[thm]{Remark}

\usepackage{enumitem}

\usepackage{caption}

\usepackage{hyperref}
\hypersetup{colorlinks=true, linkcolor=blue, urlcolor=blue}
\hypersetup{colorlinks=true, citecolor=blue}

\DeclareMathAlphabet{\mathpzc}{OT1}{pzc}{m}{it}

\usepackage{amsmath, amsthm, amssymb}

\usepackage{cleveref}

\DeclarePairedDelimiter\abs{\lvert}{\rvert}

\makeatletter
\newsavebox{\@brx}
\newcommand{\llangle}[1][]{\savebox{\@brx}{\(\m@th{#1\langle}\)}%
  \mathopen{\copy\@brx\mkern2mu\kern-0.9\wd\@brx\usebox{\@brx}}}
\newcommand{\rrangle}[1][]{\savebox{\@brx}{\(\m@th{#1\rangle}\)}%
  \mathclose{\copy\@brx\mkern2mu\kern-0.9\wd\@brx\usebox{\@brx}}}%
\let\oldabs\abs
\def\abs{\@ifstar{\oldabs}{\oldabs*}}
\makeatother

\usepackage{accents}

    {\end{bmatrix}}%

\newcommand{\armset}{\mathcal{A}}
\newcommand{\bx}{\boldsymbol{x}}
\newcommand{\bW}{\mathbf{W}}
\newcommand{\bz}{\boldsymbol{z}}
\newcommand{\regret}{\mathcal{R}}
\newcommand{\bnu}{\boldsymbol{\nu}}
\newcommand{\btheta}{\boldsymbol{\theta}}
\newcommand{\E}{\mathbb{E}}

\title{\textbf{Comparative Performance of Collaborative Bandit Algorithms: Effect of Sparsity and Exploration Intensity}}

\begin{document}

\author[]{Eren Ozbay\thanks{\texttt{	erenozzbay@gmail.com}}}
\affil[]{\small  
\textit{Discover Financial Services Inc., Riverwoods, IL 60015 USA}}
\author[]{Ashkan Golgoon\thanks{\texttt{	ashkangolgoon@gmail.com}}}
\affil[]{\small 
\textit{Independent Researcher}}
\maketitle

\begin{abstract} 
This paper offers a comprehensive analysis of collaborative bandit algorithms and provides a thorough comparison of their performance. 
Collaborative bandits aim to improve the performance of contextual bandits by introducing relationships between arms (or items), allowing effective propagation of information. Collaboration among arms allows the feedback obtained through a single user (item) to be shared across related users (items). Introducing collaboration also
alleviates the cold start problem, i.e., lack of historical information when a new user arriving to the platform with no prior record of interactions. 
In the context of modeling the relationships between arms, there are two main approaches:
Hard and soft clustering. We call approaches that model the relationship between arms in an \textit{absolute} manner as hard clustering, i.e., the relationship is binary. Soft clustering relaxes membership constraints, allowing \textit{fuzzy} assignment. 
Focusing on the latter, we provide extensive experiments on the state-of-the-art collaborative contextual bandit algorithms and investigate the effect of sparsity and how the exploration intensity acts as a correction mechanism. 
Our numerical experiments demonstrate that controlling for sparsity in collaboration improves data efficiency and performance as it better informs learning. Meanwhile, increasing the exploration intensity acts as a correction because it effectively reduces variance due to potentially misspecified relationships among users. 
We observe that this misspecification is further remedied by introducing latent factors, and thus, increasing the dimensionality of the bandit parameters. 


\end{abstract}

\begin{description}
\item[Keywords:] Recommender Systems, Contextual Bandits, Reinforcement Learning, Multi-armed Bandits.
\end{description}

\tableofcontents

\section{Introduction}\label{sec:intro}

Personalizing user experience on recommender platforms has seen growing interest from various fields, such as graph neural networks \citep{yin2019deeper,wu2022graph,gao2023survey,sharma2024survey}, reinforcement learning \citep{zou2019reinforcement,xin2020self,RLsurvey,wang2024reinforcement}, and more recently, generative models \citep{VAE_CF, neural_multiheadAttn, Seq_VAE_CF, genRecSys}, with many also focusing on their practical applications \citep{RecSysSurvey13, recsysSurvey, graphRecSys}.
The personalization task requires an online platform to effectively utilize the information it can collect on users' interactions with the platform (e.g., using their session logs) to generate insights about their preferences. 
As new information on users may emerge, or user preferences may shift, the platform needs an algorithmic approach that can quickly and reliably process new information and revise strategies accordingly. 
Hence, most platforms interacting with users implement online learning solutions to navigate this space effectively \citep{recsysRL}. 
The effectiveness of such systems relies on finding high-quality mappings (embeddings) between the users and the recommended items.

The main objective of these platforms is to ensure that users stay engaged and remain on the platform for longer, enabling higher value generation. 
Therefore, recommendations should raise users' engagement, while allowing the platform to collect new information about users' preferences. This data helps the platform refine its recommender strategies. 
This gives rise to the classic exploration-exploitation trade-off often associated with the multi-armed bandit problem in the literature. While the platform tends to optimize the information it already has about users to recommend them the best items (exploit), it also needs to capture information about users with shifting interests or emerging items (explore) to ensure that users are kept engaged. 
For example, exploration in an online advertisement setting, where ads are to be shown to users, may mean that the platform should devote a certain portion of its efforts on recommending potentially less \textit{relevant} ads to users to learn more about their preferences. This online advertisement setting will serve as our running example for the rest of the paper.

In this work, we focus on analyzing the exploration-exploitation trade-off using multi-armed bandit algorithms, which provide a principled solution for this dilemma.
Formally, we say that an arm is an item that can be recommended (assigned) to a user. The goal of the decision-maker (learner) is to recommend specific items to specific users in order to maximize the instances of users choosing to interact with the items. We assume a set of items and a static set of users with (deterministic) information known to the learner at the time of interaction and an unknown parameterized function (usually with a parameter vector $\btheta$) that maps the best item to a specific user.
The decision-maker aims to learn this function by continuously interacting with users over the course of a fixed and known time horizon, and to minimize the {\it regret}, which is the cost incurred due to uncertainty. 
Using a version of the LinUCB algorithm \cite{linucb2010}  as a baseline, we precisely evaluate more complex methods for this problem.  

Starting with the above setting of utilizing contextual information, one may imagine numerous models here: (1) use contextual information on items and associate users with bandit parameters, (2) use contextual information on users and associate items with bandit parameters, (3) combine contextual information of items and users (i.e., interactions of different contexts), and associate items or users with bandit parameters.\footnote{An algorithm for the setting of combining contextual information of items and users is presented in \cite{linucb2010} as the hybrid model.} 
We assume that there are many more users than items in our system. Hence, we seek models that relate different users' behaviors to each other, which helps to (1) reduce the amount of information kept in and fed to the model, and, (2) remedy the cold-start problem, i.e., users with no past history of interactions in the system. 
Through this relation, a new user's preferences can be modeled using the preferences of existing users.

In online advertising, the reward signal—whether or not a user clicked on an ad—is both sparse and highly variable. It is sparse because few users click on ads, and it has high variance because many of those clicks are accidental, not intentional \citep{highConfOPE}. 
Therefore, identifying unique users is crucial. This allows us to track how their preferences evolve over time, capture their underlying interests more accurately, and ultimately build long-term relationships that maximize payoff. In this model, we define the payoff as 1 if a user clicks on an ad, and 0 otherwise.
%
Note that, even though the users may be identified (which is getting harder due to privacy protections and laws, e.g., GDPR Cookie Law)\footnote{gdpr.eu/cookies}, the signals may still be sparse, and a policy interacting with new users may be inefficient if it primarily relies on user features.
For reasons mentioned above, we focus on modeling the relationship between different users to both increase the data efficiency, potentially reducing learning complexity and to facilitate the problem of cold users. 
The idea of {\it collaboration} may also improve performance with new items recommendation, and a similar approach may be utilized to map items to items to solve the cold item problem.

In order to confirm that introducing a collaboration structure indeed helps algorithms to be more data efficient, and therefore improve their performance, we conduct comprehensive experiments to compare and contrast M-LinUCB with the CoLin \citep{wu2016contextual} and FactorUCB \citep{wang2017factorization} algorithms.\footnote{M-LinUCB is a version of the LinUCB algorithm that is more suitable for our experiments \citep{wu2016contextual}.}
CoLin and FactorUCB algorithms model the relationships among users through a similarity matrix. In our experiments, we demonstrate that controlling for sparsity in this user-similarity matrix further improves data efficiency and performance as it informs relationships among users more effectively. Additionally, FactorUCB introduces latent factors to expand the dimensionality of the item feature space. We observed that adding latent factors helped FactorUCB significantly outperform CoLin. We believe this is because the latent factors compensate for a potentially inaccurate collaboration structure, which is represented by the similarity matrix. 
In order to test the effects of this potential misspecification further, we investigate the exploration intensity, and analyze the extent to which it acts as a correction. We argue that increasing the exploration intensity helps as it balances the variance due to misspecification through putting more emphasis on exploratory actions. Note that this behavior may not be consistent across different paradigms where underlying distributions quickly stabilize.



We use the Yahoo! Today Module (2009) dataset (version \textup{R6A}) in our experiments \citep{linucb2010}.\footnote{Obtained from webscope.sandbox.yahoo.com/catalog.php?datatype=r}
This dataset is considerably large with over $45$ million data points, $20$ candidate articles, and a total of $270$ articles over the course of a ten-day period. 
The payoff of the interaction of the user with the recommended article is called a {\it click}, as described earlier, and is $1$ if the user clicks on the shown article, and $0$ otherwise. 
Also note that, because this dataset is de-identified, we cannot treat users independently and we cluster users before running any of the algorithms we analyze. 

As the user-article interaction data is very sparse in the Yahoo! dataset, we observe the importance of using data efficiently in learning user preferences. For example, when we increase the number of clusters we group users into, M-LinUCB performs worse. This is because increasing the number of clusters results in increased sparsity for each user group, and because M-LinUCB does not share collected feedback across distinct clusters, it performs worse compared to a smaller cluster number.

While a collaboration structure helps algorithms to improve their performance, it comes at a considerable computational cost. CoLin \citep{wu2016contextual}, with 10 times computational cost compared to that of M-LinUCB, enjoys an improvement of only $7\%$. On the other hand, FactorUCB \citep{wang2017factorization}, an extension of CoLin, enjoys an improvement of $33\%$ over M-LinUCB, with 20 times computational cost compared to that of M-LinUCB. FactorUCB's improved performance is due to introducing latent factors. 


The rest of the paper is organized as follows: In Section \ref{sec:lit}, we survey the relevant works. 
In Section \ref{sec:prel}, we present the preliminaries and the algorithms we use in our experiments. 
We extensively analyze the results in Section \ref{sec:numerics} and Section \ref{sec:concl} discusses our findings and proposed future directions, both empirical and technical, in detail.

\section{Related Work}\label{sec:lit}
We examine two principal types of algorithms in this field, namely offline and online learning algorithms. We categorize collaborative filtering and matrix factorization based algorithms as offline (though we refer the reader to some online applications of these algorithms \citep{zangerle2022evaluating,raza2022news,chen2023deep}). Then, we survey the collaboration-enabled algorithms under online learning. This collaboration is generally done through (weighted) graphs. Here, we consider clustering as a graph-based model.

\subsection{Offline Setting: Collaborative Filtering and Matrix Factorization}
Collaborative Filtering (CF) \citep{CF_orig} is one of the more popular approaches in recommender systems, mostly due to its data efficiency. CF either uses (1) a neighborhood-based model to find relationships between items or between users and make recommendations based on this shared knowledge graph to relate items to items, or, (2) a latent factor model, e.g., Matrix Factorization (MF), to embed items and users to the same latent factor space and compare both in that space. 

\cite{koren2008factorization} notes that neighborhood based models are usually more effective at detecting local relationships, failing to capture weaker relationships that may still help in recovering the underlying similarities \citep{itemCF, MF_RecSys}. 
On the other hand, latent factor models are usually more effective at capturing underlying similarities that may be less apparent at a lower dimensional space.
As the embedding procedure targets a more generalized model, they put less emphasis on local yet strong relationships \citep{Prob_MF, hu2008collaborative, poisson_factor, ALS_survey}. 
Also note that, \cite{hu2008collaborative} shows their MF-based model can be reduced to an item-item neighborhood model, suggesting that there are certain modeling approaches where these two model classes may be mapped to each other.

In extending to more state-of-the-art methods in the literature, a growing recent body of work involves introducing non-linearity into traditional CF and MF methods. 
Many works use neural networks \citep{deepCF, CDAE, neuralautoregressiveapproachcollaborative, he2017neural, RNN_rec, NeuralGraphCF, LightGCN, DisentangledGraphCF, wu2022graph}, with 
\cite{AutoRec} extending autoencoders \citep{autoencoder}, and
\cite{VAE_CF} extending variational autoencoders \citep{autoencodingvariationalbayes}.
\cite{BERT4Rec} use the transformer structure \citep{BERT} to exploit the time effect and optimize the learning through user-behavior sequences.



While generally implemented offline, both CF and MF models, can build a time series model and track the shifts in user preferences, item characteristics, and cross relationship between users and items; see \citep{Koren2022} for a detailed survey on the state-of-the-art for time-dependent CF and MF methods and \citep{BERT4Rec} for a transformer based approach. 
While building a time-dependent model does capture the shift in user preferences, it is not truly online in the sense that a deployed model does not necessarily update its parameters with newly arriving data. Although there are online implementations of CF and MF \citep{online_CF, he2016fast}, we preferentially call the multi-armed bandit algorithms in the next section as {\it online}.

\subsection{Online Setting: Collaboration in Bandits}
Contextual bandit model \citep{linucb2010}, different than the traditional bandit problems, assumes that the expected payoff of an action is a linear combination of the contextual information and an arm-specific bandit parameter that is initially unknown and learned through sequential experimentation. Similar to neighborhood based or latent factor models, recent contextual bandit models incorporate sharing of information across users or items, and use latent factors to generalize the linear model. 

We focus on three main works: LinUCB algorithm \citep{linucb2010}, CoLin algorithm \citep{wu2016contextual}, and FactorUCB algorithm \citep{wang2017factorization}. The LinUCB algorithm introduces the use of contextual bandit modeling. CoLin builds onto LinUCB by introducing user neighborhoods to relate users to each other. Furthermore, FactorUCB builds onto CoLin by incorporating latent factors. In our experiments, we analyze and comment on this evolution of models, providing additional context on each of these algorithms and how they achieve the observed improvements.

Note that the above algorithms attempt to model the dependency among users through a weighted adjacency graph an approach known as soft-clustering. On the other hand, there is another class of algorithms that focus on hard-clustering (i.e., binary assignment of classes) and requiring the learned bandit parameters to be kept homogeneous across all users in the same cluster. Next, we review some of these approaches that utilize hard-clustering method to group items (or users).

\cite{cesa2013gang} require users connected in a neighborhood graph to have similar bandit parameters, explicitly forcing the learned bandit parameters to be close to each other during learning. This improves upon LinUCB as it uses relational information between users to share knowledge gathered about the interests of a given user to their {\it friends}. Exploiting such a network structure, the authors show a consistent increase in prediction performance. 
\cite{gentile2014online} also use an approach that clusters the users, which force the users in the same cluster to share the same bandit parameters. They start with a complete graph, i.e., all users are connected to each other, and remove connections they deem weak. This improves upon \citep{cesa2013gang} in instances where cluster information is not readily available, or where clusters defined (and fixed for the remainder of the time horizon) using the information obtained so far is a weak representation for the underlying similarity structure.

Building on \citep{gentile2014online}, \cite{gentile2017context} 
cluster users based on items and condition on their interactions with the items being the same. In other words, if two users $j$ and $k$ are in the same cluster from the viewpoint of item $\bx$ and at time $t$, then it must be that $\theta_{j, t} \bx = \theta_{k, t} \bx$, where $\theta_{i, t}$ is the bandit parameter estimate for user $i$ at time $t$. Then, the reward signal is propagated across users in the same neighborhood as if all such users have interacted with the item, conditioned on some threshold. This allows users to be added to a new neighborhood or dropped from it on the fly. This is in contrast to \cite{gentile2014online}, which only removes connections and does not create any new connections. Note that, however, allowing clusters to be updated across items may make the algorithm less robust and create high variance, particularly in instances of new users and items.

Similarly, \cite{li2016online} borrow item neighborhoods (item-item similarities) idea from CF to combine clustering of users with clustering of items to further remedy cold-start problem involving items. Moreover, \cite{li2016collaborative} recognize that the clusters of users (or items) can be item (or user) dependent, but their work is limited by the assumption that the content universe has to be finite and known in advance.

Extending on \citep{gentile2017context}, \cite{li2019improved} aim to solve the non-uniformity problem in users' interactions with the items. The non-uniformity arises because there may be a small set of users who interact with the platform the most and this imbalance in interaction frequencies should be utilized in splitting or combining clusters.

\section{Problem Formulation}\label{sec:prel}
We first review fundamentals of contextual bandit algorithms.
We describe the preliminaries on contextual bandits and set the notation for the remainder of the paper. Next, we introduce the collaboration structure into contextual bandit algorithms and present the three algorithms relevant to this work: LinUCB \citep{linucb2010}, CoLin \citep{wu2016contextual}, and FactorUCB \citep{wang2017factorization}.

We consider a stochastic contextual multi-armed bandit problem with a finite number of arms belonging to an arm set $\armset$, where $|\armset| = K$, for some known $K$. We use $\armset_t$ to denote the set of arms available at time $t$, under the condition that $|\armset|$ does not grow prohibitively large and arms in $\armset_t$ are distinguishable. The length of the time horizon (i.e., number of interactions or {\it pulls}) is denoted by $T$ and the number of users is denoted by $N$. We assume that the set of users is static throughout the time horizon $T$.

At each time $t \in [T]$, a user $u_t$ arrives to the system with context $u_t(x) \in \mathbb{R}^{d_u}$.\footnote{$[n]$ denotes the natural numbers up to and including $n$, i.e., $[n] = \{1,2,\dots,n\}$.} 
Context of each arm in $\armset$ at time $t$, $a_t(x) \in \mathbb{R}^{d_a}$, may be combined with context of user $u_t$ to obtain $\bx_{a_t, u_t} \in \mathbb{R}^d$, where $d = d_u + d_a$.\footnote{Note that the context of the user may change across time, e.g., number of recommendations they received so far may be one of the information dimensions \citep{LTV}. We assume that $d_u$ and $d_a$ remain unchanged throughout the time horizon.} 
After the decision-maker (learner) pulls arm $a_t$ at time $t$, 
they receive a payoff of $r_{a_t, u_t}$ from user $u_t$. This payoff is assumed to be a function of the context and the system state at time $t$, i.e., $r_{a_t, u_t} = f(\bx_{a_t, u_t}, \boldsymbol{S}_t)$, where $\boldsymbol{S}_t$ may contain any 
information collected up to time $t$, and $f$ is an arbitrary function. Note that the learner has access to $\boldsymbol{S}_0$ prior to any interactions.

We further assume that the expected payoff at time $t$ can be 
represented 
as the dot product of the context and an unknown vector\footnote{We make the (linear) {\it realizability} assumption that such a vector exists \citep{auer2002, linucb_ext2011}. This implies that, given perfect information, i.e., the vector $\boldsymbol{\theta}_a^*$, the learner would always pick the optimal arm at time $t$.}, i.e.,
\begin{align}
    \E\left[ r_{a_t,u_t} \mid \bx_{a_t, u_t} \right] = \bx_{a_t, u_t}^{\textup{T}} \btheta_{a_t}^*, \label{eqn:payoff}
\end{align}
where
$\btheta_{a_t}^*$ is a time-invariant 
and unknown vector (bandit parameter) associated with arm $a_t$. 

Let $\mathcal{H}_t$ denote the history of actions and observations at time $t\geq2$ with 
$\mathcal{H}_t = \left(a_n, r_{a_n,u_n}, \armset_n, \boldsymbol{S}_n \right)_{n \in [t-1]}$ and $\mathcal{H}_1 = (\armset_0, \boldsymbol{S}_0)$. 
Then, a policy $\pi$ of the learner is a sequence of mappings $(\pi_1, \pi_2, \dots, \pi_T)$ where $\pi_t: \mathcal{H}_t \mapsto a_t$ for $t \in [T]$. Let $\Pi_T$ denote the set of all such policy $\pi$.

Once a policy is fixed by the learner, the actions and observed rewards become well-defined random variables for all time periods $t \in [T]$. Because of our earlier modeling assumption in Equation~\ref{eqn:payoff}, we define the $T$-period regret of the learner with respect to an oracle as follows:
\begin{align}
\regret_T(\pi, \bnu) = \E\left[ \sum_{t=1}^T r_{a_t^*,u_t} \right] - \E\left[ \sum_{t=1}^T r_{a_t,u_t} \right] \label{eqn:regret},
\end{align}
where $a_t^*$ is the arm chosen by an oracle at time $t$ and $\bnu$ is the true distribution of all bandits, i.e., $\btheta_a^* \sim \nu_a$, $a \in \armset$. Note that any oracle has access to $\bnu$.

Then, the objective of the learner is to design a policy $\pi_T \in \Pi_T$ such that the smallest regret in the worst-case over all distributions $\bnu \in \mathcal{V}$ is achieved, i.e., 
\begin{align}
     \inf_{\pi \in \Pi_T} \sup_{\bnu \in \mathcal{V}} \regret_T(\pi, \bnu),
\end{align}
where $\mathcal{V}$ is the set of all $K$-tuples of distributions for $K$ arms having support in $[0,1]$.


Contextual bandit models provide a framework to incorporate the abundant information collected from users and items (i.e., arms) into decision making. 
In terms of interacting with users, by taking an action (such as recommending an item) and capturing the user's feedback, most algorithms work in a similar manner. What distinguishes these algorithms is how they use the collected information, and how they share it among all users and actions. To materialize this, we introduce Algorithm~\ref{alg:sketch} as a basis for all contextual bandit algorithms.

\begin{algorithm}[h]
	\caption{A Basic Contextual Bandit Algorithm}\label{alg:sketch}
	\vspace{0.1cm} 
	{\bf Input:} Set of parameters $\mathcal{C}$, set of arms $\armset_t$ for $t \in [T]$, initial system information $\boldsymbol{S}_0$, function $g(\cdot)$\\
	\For{$t \in [T]$}
        {
            Receive user $u_t$ and observe $\bx_{a_t, u_t}$ $\forall a \in \armset_t$\\
            Pull arm $a_t$ that maximizes $g(\bx_{a_t, u_t}, \boldsymbol{S}_t)$ and observe reward $r_{a_t, u_t}$\\
            Update the parameters and the system information 
        }
\vspace{0.1cm}
\end{algorithm}

The function $g(\cdot)$ in line 4 of Algorithm \ref{alg:sketch} can be a function that is linear in context and system information (as in LinUCB \citep{linucb2010}), or a {\it link} function that is non-linear in its arguments \citep{GLMbandit}. Note that, while we primarily focus on linear models, some offline models, such as the celebrated XGBoost \citep{chen2016xgboost}, are non-linear. 

We further focus on a subset of linear models, namely upper confidence bound (UCB) based methods, to define the function $g(\cdot)$ in line 4 of Algorithm \ref{alg:sketch}.\footnote{We abstain from arguing for the efficacy of UCB-based methods and refer the reader to the seminal papers on UCB1 \citep{auer2002finite, linucb2010}.} This function consists of two elements: (1) The predicted payoff of the action based on information gathered so far, and (2) an {\it error} term that ensures the first term is not too far away from the true expected payoff with high probability. This second term can be thought of as the {\it standard deviation}, which when multiplied by a large enough constant, guarantees that the summation of both terms will overestimate the true expected payoff with high probability. We call the arm that has the highest UCB at time $t$ as the empirically {\it best} arm. 


\subsection{Contextual bandits in collaboration} 


We first introduce the LinUCB in Algorithm~\ref{alg:linucb} as our base algorithm. Then, we define what {\it collaboration} entails and introduce CoLin \citep{wu2016contextual}, a collaborative contextual bandit algorithm, by modifying Algorithm~\ref{alg:linucb}.

\begin{algorithm}[h]
	\caption{LinUCB \citep{linucb2010}}\label{alg:linucb}
	\vspace{0.1cm} 
	{\bf Input:} $\alpha \in \mathbb{R}_+$, initial arm set $\armset_0$\\
        \For{$t \in [T]$}
        {   
            Receive user $u_t$ and observe $\bx_{a, u_t}$ $\forall a \in \armset_t$\\
            \For{{\bf all} $a \in \armset_t$}
            {
                \If{a is new}
                {
                    $\mathbf{A}_a \gets \mathbf{I}_{d\times d}$\\
                    $\mathbf{b}_a \gets \mathbf{0}_{d\times 1}$\\
                }
                $\hat{\boldsymbol{\theta}}_a \gets \mathbf{A}_a^{-1} \mathbf{b}_a$\\
                $p_{t,a} \gets \hat{\boldsymbol{\theta}}_a^{\textup{T}} \bx_{a, u_t} + \alpha \sqrt{\bx_{a, u_t}^{\textup{T}} \mathbf{A}_a^{-1} \bx_{a, u_t}}$\\
            }
            Pull arm $a_t = \arg\max_{a \in \armset_t} p_{t,a}$ and observe reward $r_{a_t, u_t}$\\
            $\mathbf{A}_{a_t} \gets \mathbf{A}_{a_t} + \bx_{a_t, u_t} \bx_{a_t, u_t}^{\textup{T}}$\\
            $\mathbf{b}_{a_t} \gets \mathbf{b}_{a_t} + r_{a_t, u_t} \bx_{a_t, u_t}$\\
        }
\vspace{0.1cm}
\end{algorithm}

Consider Algorithm~\ref{alg:linucb} and notice that lines 4--12 and 13--14 correspond to finding and pulling the {\it best} arm and updating the model parameters, respectively. Here, we let 
\[
    g(\bx_{a_t, u_t}, \boldsymbol{S}_t) \triangleq p_{t,a} = \hat{\boldsymbol{\theta}}_{a_t}^{\textup{T}} \bx_{a_t, u_t} + \alpha \sqrt{\bx_{a_t, u_t}^{\textup{T}} \mathbf{A}_{a_t}^{-1} \bx_{a_t, u_t}},
\]
where $\hat{\boldsymbol{\theta}}_a$ is the current estimate for the true bandit parameter of arm $a_t$ at time $t \in [T]$.

\begin{remark} Algorithm~\ref{alg:linucb} is extended in \cite{linucb2010} to include a certain level of collaboration across all arms. 
Focusing only on the updated $g(\cdot)$ function, which also includes a bandit parameter shared by all arms:
\[
    g(\bx_{a_t, u_t}, \bz_{a_t, u_t}, \boldsymbol{S}_t) \triangleq p_{t,a} = \hat{\boldsymbol{\beta}}_t^{\textup{T}} \bz_{a_t, u_t} + \hat{\boldsymbol{\theta}}_{a_t}^{\textup{T}} \bx_{a_t, u_t} + \alpha \sqrt{s_{a_t, u_t}},
\]
where $\hat{\boldsymbol{\theta}}_a$ is the current estimate for the true bandit parameter of arm $a_t$ at time $t \in [T]$ and $\hat{\boldsymbol{\beta}}_t$ is the current estimate for the true bandit parameter shared by all arms.
Here, $s_{a_t, u_t}$ is a more involved version of the {\it standard deviation} term of $g(\cdot)$ function in Algorithm \ref{alg:linucb}. 
The vectors 
$\bz_{a_t, u_t}$ and $\bx_{a_t, u_t}$ both contain contextual information: $\bx_{a_t, u_t}$ may be a concatenation of user and arm contexts, whereas $\bz_{a_t, u_t}$ is a product of user and arm contexts and $\bz_{a_t, u_t}\in \mathbb{R}^{d_u \cdot d_a}$.
Note that in this setting we have the expected payoff at time $t \in [T]$
\[
    \E\left[ r_{a_t,u_t} \mid \bx_{a_t, u_t}, \bz_{a_t, u_t} \right] = \bz_{a_t, u_t}^{\textup{T}} \boldsymbol{\beta}^* + \bx_{a_t, u_t}^{\textup{T}} \boldsymbol{\theta}_{a_t}^*,
\]
where $\boldsymbol{\theta}_{a_t}^*$ is the true bandit parameter of arm $a_t \in \armset_t$ and $\boldsymbol{\beta}^*$ is the true bandit parameter shared by all arms.
The interactions of users with arms other than arm $a$ can now affect the learning of the true bandit parameter of arm $a$ to a certain extent. The product context vector $\bz$ also allows for highlighting interactions of different contextual dimensions \citep{linucb2010}. This is a model to inject collaboration across arms.
\end{remark}


\begin{algorithm}[h]
    \caption{M-LinUCB \citep{linucb2010}}\label{alg:m-linucb}
    \vspace{0.1cm} 
    {\bf Input:} $\alpha \in \mathbb{R}_+$, initial item set $\armset_0$, user cluster set $\mathrm{C}  \triangleq \cup_{i=1}^M \mathrm{C}_i$.\\
    {\bf Initialize:} $\mathbf{A}_{\mathrm{C}_i} \gets \mathbf{I}_{d\times d}$, $\mathbf{b}_{\mathrm{C}_i} \gets \mathbf{0}_{d\times 1}$ $\forall i \in [M]$.\\
    \For{$t \in [T]$}
    {   
        Receive user $u_t$\\
        \If{$u_t$ is new}
        {
            Assign $u_t$ to a cluster
        }
        $i \gets \left\{i : u_t \in \mathrm{C}_i \right\}$\\
        $\hat{\boldsymbol{\theta}}_{\mathrm{C}_i} \gets \mathbf{A}_{\mathrm{C}_i}^{-1} \mathbf{b}_{\mathrm{C}_i}$\\
        \For{{\bf all} $a \in \armset_t$}
        {
            Observe $\bx_{a, u_t}$ \\
            $p_{t,a} \gets \hat{\boldsymbol{\theta}}_{\mathrm{C}_i}^{\textup{T}} \bx_{a, u_t} + \alpha \sqrt{\bx_{a, u_t}^{\textup{T}} \mathbf{A}_{\mathrm{C}_i}^{-1} \bx_{a, u_t}}$\\
        }
        Pull arm $a_t = \arg\max_{a \in \armset_t} p_{t,a}$ and observe reward $r_{a_t, u_t}$\\
        $\mathbf{A}_{\mathrm{C}_i} \gets \mathbf{A}_{\mathrm{C}_i} + \bx_{a_t, u_t} \bx_{a_t, u_t}^{\textup{T}}$\\
        $\mathbf{b}_{\mathrm{C}_i} \gets \mathbf{b}_{\mathrm{C}_i} + r_{a_t, u_t} \bx_{a_t, u_t}$\\
    }
\vspace{0.1cm}
\end{algorithm}

Notice that we associate arms with bandit parameters in Algorithm \ref{alg:linucb}. 
To extend this to a setting where arms are more persistent than users, we instead associate users with bandit parameters and use the contextual information on arms to design the best policy.
Additionally, we consider a setting where there are prohibitively many users and we remedy this by clustering users and associating these clusters with bandit parameters. 
In this setting, whenever a user belonging to a specific cluster gets an arm assigned to them, it is as though all the other users in the same cluster effectively do so. 
This also helps in propagating the information collected on a user to another one based on their perceived similarity. Assuming that a structured way to extract such similarity information exists, such as already existing friend networks in social media, collaboration propagates new information and thus speeds up the learning process, and also offers a remedy to the cold-start problem in recommender systems. The intuition is similar to that in \cite{linucb2010}: The payoff from a user’s interaction may be a function of both their preference and the influence of others.

There are two main approaches in the literature to inject collaboration into contextual bandit models, namely hard- and soft-clustering. 
In hard-clustering, connected arms are assumed to have similar or exactly the same bandit parameters. For instance, GOB.Lin \citep{cesa2013gang} requires connected users in a network to share similar bandit parameters through a graph Laplacian-based regularization, i.e., the bandit parameters across connected users must be close to each other in space. CLUB \citep{gentile2014online} makes a similar but stronger modeling assumption and forces arms in the same cluster to have exactly the same bandit parameters.
In soft-clustering, arms are {\it connected} through a weighted adjacency graph, where nodes are contextual bandits on users and the weights on the edges govern the influence between pairs of users. In upcoming parts, we introduce the notion of {\it sparsity} in these settings. We define 100\% sparsity as the state where the weighted adjacency graph keeps all weights non-zero.
A $50\%$ sparsity then means that the weighted adjacency graph keeps only the largest half of the weights non-zero.\footnote{This is from each user's perspective, i.e., this is a directed graph and not necessarily symmetric in both directions between nodes.}


\begin{algorithm}[h]
    \caption{CoLin \citep{wu2016contextual}}\label{alg:colin}
    \vspace{0.1cm} 
    {\bf Input:} $\alpha \in \mathbb{R}_+$, $\armset_0$, $\bW \in \mathbb{R}^{M \times M}$. \\
    {\bf Initialize:} 
    $\mathbf{A}_1 \gets \mathbf{I}_{dM \times dM}$, $\mathbf{b}_1 \gets \mathbf{0}_{dM \times 1}$, 
    $\mathbf{E}_1 \gets (\bW^{\textup{T}} \otimes \mathbf{I}_{d \times d}) \mathbf{A}_1^{-1} (\bW \otimes \mathbf{I}_{d \times d}), \widehat{\Theta}_1 \gets \mathbf{0}_{d \times M}.$\\
    \For{$t \in [T]$}
    {   
        Receive user $u_t$\\
        \If{$u_t$ is new}
        {
            Assign $u_t$ to a cluster
        }
        $i \gets \left\{i : u_t \in \mathrm{C}_i \right\}$\\
        \For{{\bf all} $a \in \armset_t$}
        {
            Observe $\bx_{a, u_t} \in \mathbb{R}^{d\times 1}$\\
            $p_{t,a} \gets \left[(\widehat{\Theta}_t \bW)^{\textup{T}}\right]_{(i)} \bx_{a, u_t} + 
            \alpha \sqrt{ \tilde{\bx}_{a, u_t}^{\textup{T}} \mathbf{E}_t \tilde{\bx}_{a, u_t} } $\\
        }
        Pull arm $a_t = \arg\max_{a \in \armset_t} p_{t,a}$ and observe reward $r_{a_t, u_t}$\\
        $\mathbf{A}_{t+1} \gets \mathbf{A}_t + \textup{h}(\tilde{\bx}_{a_t, u_t}\otimes_{\textup{outer}}[\bW^{\textup{T}}]_{(i)})$\\ 
        $\mathbf{b}_{t+1} \gets \mathbf{b}_t + r_{a_t, u_t} (\tilde{\bx}_{a_t, u_t}\otimes_{\textup{outer}}[\bW^{\textup{T}}]_{(i)})$\\
        $\mathbf{E}_{t+1} \gets (\bW^{\textup{T}} \otimes \mathbf{I}) \mathbf{A}_{t+1}^{-1} (\bW \otimes \mathbf{I})$\\
        $\widehat{\Theta}_{t+1} \gets \textup{mat}_{d,M}(\mathbf{A}_{t+1}^{-1} \mathbf{b}_{t+1})$\\
    }
\vspace{0.1cm}
\end{algorithm}

We can introduce hard-clustering to LinUCB in Algorithm~\ref{alg:linucb} and comment on how to inject collaboration by modifying it to obtain M-LinUCB in Algorithm~\ref{alg:m-linucb}. 
Notice that in Algorithm~\ref{alg:m-linucb}, we associate users with bandit parameters and assume all users, new or existing at time $t=0$, belong to a cluster (deterministically). In Algorithm~\ref{alg:m-linucb}, we have $M$ user clusters, and each user $u$ belongs to one of those clusters. 
We do not model any dependencies among clusters in Algorithm~\ref{alg:m-linucb} and therefore no information is propagated across clusters.

We expect to observe performance improvements over Algorithm~\ref{alg:m-linucb} if we model the potential similarity between clusters, or forgo the hard-cluster structure and introduce a weighted adjacency graph to inject collaboration. 
Therefore, we extend M-LinUCB to obtain CoLin \citep{wu2016contextual}, and extend CoLin further by including latent features to it, obtaining FactorUCB \citep{wang2017factorization}.

As clusters are intertwined in CoLin, that is, the collaboration structure allows users to influence other users, we can no longer let model parameters be cluster-specific without introducing further unnecessary complexity. 
Hence, we time-index the parameters, e.g., $\mathbf{A}_{\mathrm{C}_i}$ (per cluster) becomes $\mathbf{A}_t$ (per time). This is enabled through introducing a block structure to permit cross relationships between items and users.

Before we introduce CoLin and FactorUCB, we define three special functions: $\textup{h}(\cdot)$ is the self-outer-product function, i.e., $\textup{h}(x) = x \otimes_{\textup{outer}} x$. Note that $\otimes$ is the standard Kronecker product for matrices which outputs a block matrix. The operation $\textup{mat}_{r,c}(\mathbf{x})$ on the vector $\mathbf{x} \in \mathbb{R}^{rc \times 1}$ outputs an $r \times c$ matrix. 
$\tilde{x}$ pads a row vector $x$ with preceding and/or trailing zeros as needed.\footnote{In Algorithms \ref{alg:colin} and \ref{alg:factorucb}, $\tilde{\bx}_{a, u}$ is a row vector of dimension $1\times dM$, obtained by adding $d$-many preceding and trailing zeros for each user before and after user $u_t$, respectively ($(d-1)M$ zeros are added in total).} 

\begin{algorithm}[h]
	\caption{FactorUCB \citep{wang2017factorization}}\label{alg:factorucb}
	\vspace{0.1cm} 
        {\bf Input:} $\armset_0$, $\bW \in \mathbb{R}^{M \times M}$, $\alpha_1=\alpha_2=0.375$. \\
    {\bf Initialize:} 
    $\mathbf{A}_1 \gets \mathbf{I}_{(d+d_l)M \times (d+d_l)M}$, 
    $\mathbf{b}_1 \gets \mathbf{0}_{(d+d_l)M \times 1}$, 
    $\widehat{\Theta}_1 \gets \mathbf{0}_{(d+d_l) \times M}$, 
    $(\mathbf{E}_{a, 1} \gets \mathbf{I}_{d_l \times d_l})_{a \in \armset_0}$, 
    $(\mathbf{d}_{a, 1} \gets \mathbf{0}_{d_l \times 1})_{a \in \armset_0}$, 
    $(\mathbf{\hat{v}}_{a, 1} \gets \mathbf{0}_{d_l \times 1})_{a \in \armset_0}$.\\
    \For{$t \in [T]$}
    {   
        Receive user $u_t$\\
        \If{$u_t$ is new}
        {
            Assign $u_t$ to a cluster
        }
        $i \gets \left\{i : u_t \in \mathrm{C}_i \right\}$\\
        \For{{\bf all} $a \in \armset_t$}
        {
            Observe $\bx_{a, u_t} \in \mathbb{R}^{d\times 1}$\\
            $p_{t,a} \gets \left[(\widehat{\Theta}_t \bW)^{\textup{T}}\right]_{(i)} \left(\bx_{a, u_t}, \mathbf{\hat{v}}_{a, t} \right) + 
            \alpha_1 \sqrt{ \left(\left(\tilde{\bx}_{a, u_t}, \tilde{\mathbf{v}}_{a, t}\right) \bW^{\textup{T}}\right) \mathbf{A}_t^{-1} \left(\left(\tilde{\bx}_{a, u_t}, \tilde{\mathbf{v}}_{a, t}\right) \bW^{\textup{T}}\right)^{\textup{T}} } +
            \alpha_2 \sqrt{ \left[(\widehat{\Theta}_t^{\mathbf{\hat{v}}} \bW)^{\textup{T}}\right]_{(i)} \mathbf{E}_{a, t}^{-1} \left(\left[(\widehat{\Theta}_t^{\mathbf{\hat{v}}} \bW)^{\textup{T}}\right]_{(i)}\right)^{\textup{T}} }$\\
        }
        Pull arm $a_t = \arg\max_{a \in \armset_t} p_{t,a}$ and observe reward $r_{a_t, u_t}$\\
        $\mathbf{A}_{t+1} \gets \mathbf{A}_t + \textup{h}(\left(\tilde{\bx}_{a, u_t}, \tilde{\mathbf{v}}_{a, t}\right)\otimes_{\textup{outer}}[\bW^{\textup{T}}]_{(i)})$\\ 
        $\mathbf{b}_{t+1} \gets \mathbf{b}_t + r_{a_t, u_t} (\left(\tilde{\bx}_{a, u_t}, \tilde{\mathbf{v}}_{a, t}\right)\otimes_{\textup{outer}}[\bW^{\textup{T}}]_{(i)})$\\
        $\widehat{\Theta}_{t+1} \gets \textup{mat}_{d+d_l,M}(\mathbf{A}_{t+1}^{-1} \mathbf{b}_{t+1})$\\
        $\mathbf{E}_{a_t, t+1} \gets \mathbf{E}_{a_t, t} + \textup{h}([\widehat{\Theta}_t^{\mathbf{\hat{v}}} \bW]_{(i)})$\\
        $\mathbf{d}_{a_t, t+1} \gets \mathbf{d}_{a_t, t} + ([\widehat{\Theta}_t^{\mathbf{\hat{v}}} \bW]_{(i)}) (r_{a_t, u_t} - \bx_{a_t, u_t}^{\textup{T}} [\widehat{\Theta}_t^{\bx} \bW]_{(i)})$\\
        $\mathbf{\hat{v}}_{a, t+1} \gets \mathbf{E}_{a_t, t+1}^{-1} \mathbf{d}_{a_t, t+1}$
    }
\end{algorithm}

We present CoLin in Algorithm \ref{alg:colin} and FactorUCB in Algorithm \ref{alg:factorucb}.
In both algorithms, we restrict the number of items available in each time period, i.e., $|\armset_t| = K$ $\forall t \in [T]$, for some known and time-invariant $K$. 

Comparing M-LinUCB in Algorithm \ref{alg:m-linucb} to CoLin in Algorithm \ref{alg:colin}, we observe the introduction of the so-called similarity matrix $\bW$. This matrix relates the estimated bandit parameter of each user cluster to that of other user clusters. An element $(i,j)$ of the similarity matrix $\bW$ governs how much the bandit parameter associated with user cluster $i$ affects the bandit parameter associated with user cluster $j$. That is, column $j$ of $\bW$ contains all the information necessary to propagate the user preferences from all other clusters $i \neq j$. The similarity matrix $\bW$ is column-normalized 
and non-negative. This implies that a bandit parameter for a cluster is defined as the convex combination of all clusters' bandit parameters, including its own. 

Observe that in Line 12 of Algorithm \ref{alg:m-linucb}, the UCB of an article is calculated by only using the bandit parameter for the current user's cluster. In Line 11 of Algorithm \ref{alg:colin}, UCB is now calculated by generating a {\it shared} bandit parameter for all user clusters the current user has related preferences with. Similarly, the {\it standard deviation} term in Line 11 of Algorithm \ref{alg:colin} now depends on the similarity matrix $\bW$ as well (compare to that in Line 12 of Algorithm \ref{alg:m-linucb}).

Then, going from CoLin to FactorUCB in Algorithm \ref{alg:factorucb}, we observe two new concepts: (1) increase in the dimension of the bandit parameters by the dimension of latent factors ($d_l$) introduced to the items, increasing the dimensionality to $d + d_l$, and (2) introduce a learning scheme for latent item factors that make up the context for the items, i.e., item contexts are made up of two parts: static and dynamic. Although we combine the unknown parameters in $\widehat{\Theta}$, in calculating user specific bandit parameters and arm specific latent factors, we let $\widehat{\Theta}^{\bx}$ denote the user part and $\widehat{\Theta}^{\mathbf{\hat{v}}}$ denote the arm (latent factor) part, i.e., $\widehat{\Theta} = \textup{concat}\left( \widehat{\Theta}^{\bx}, \widehat{\Theta}^{\mathbf{\hat{v}}}\right)$.\footnote{We use $\hat{q}$ to denote that $q$ is a learnable parameter.} In the statement of the algorithm, we drop ``$\textup{concat}$" and simply put two vectors within closed parentheses to represent that they are concatenated.

Lastly, before moving on to the experiments, we note that
in all experiments, the same similarity matrix $\bW$ is used for CoLin and FactorUCB for set values of cluster sizes and sparsity values. And, 
we use user features in finding the cluster centers and assigning users to cluster, and only use the known article features in $\bx$.



\section{Numerical Experiments}
\label{sec:numerics}

We compare and contrast M-LinUCB (Algorithm~\ref{alg:m-linucb}), CoLin (Algorithm~\ref{alg:colin}) and FactorUCB (Algorithm~\ref{alg:factorucb}) algorithms on the Yahoo! Today Module (2009) dataset (version \textup{R6A}).\footnote{Obtained from \url{https://webscope.sandbox.yahoo.com/catalog.php?datatype=r&guccounter=1}} First, we detail our experiments, then we discuss how varying sparsity and exploration intensity affect the results.

We associate bandit parameters with users because we aim to model the similarity across users' preferences rather than items' characteristics. 
Moreover, the choice of associating bandit parameters with users is often more convenient as the items' characteristics are under the control of the decision-maker, while the users' characteristics are exogenous to the platform and harder to account for. Note that, in general, many considerations may affect this choice, e.g., number of items or users and their ratios.



\subsection{Offline Policy Evaluation and Sparsity}
Before proceeding with the details of our numerical experiments, we discuss the policy evaluation method that we adapt, and how collaboration is handled for the relevant algorithms.
\vspace{0.2cm}

{\bf Policy evaluation method. }
The sequence of events for the evaluation of the Yahoo! dataset are as follows:
\begin{enumerate}
    \item Receive a user 
    and a pool of candidate arms the user can receive as recommendation. For 
    the Yahoo! Today Module dataset, this pool consists of 20 articles that are shown in an instance.
    \item Obtain the recommended article selected by the policy and check if it matches that of the historical data. If the recommendation matches the historical data, record the reward and update the model parameters. Otherwise, do not update the model parameters.\footnote{This choice of policy evaluation is selected for the sake of simplicity. In general, one can use more advanced offline policy evaluation methods, such as the doubly robust method \citep{OPE_DR}.}
\end{enumerate}

We utilize user features to create cluster centroids, assigning each user to the nearest cluster. We then associated these clusters with the bandit parameters. We use the cluster centroids to define the collaboration across clusters. The dot product of the centroid of cluster $i$ with that of cluster $j$ is used to obtain the intensity of the collaboration. We then normalize these values to generate the collaboration intensity and call them {\it weights}, each of which is a number between $0$ and $1$. 

We also control the sparsity in this collaboration structure. By controlling sparsity, we mean only preserving the top $x\%$ weights across users. In this case, we say that this instance has Sparsity $x\%$. If we introduce sparsity (i.e., set $x$ to a positive number strictly less than $100$), we normalize the weights before using them in the algorithms.


\subsection{Results on Yahoo! Today Module Dataset}
\label{sec:exp_yahoo}

One of the immediate problems with the Yahoo! Today Module (2009) dataset is that there are no unique users that can be identified. This is due to the feature reduction step performed by the owners of the dataset to mask user identities (see, e.g., \citep{linucb2010} for details). This fact makes it necessary to cluster users into groups. 
Therefore, we perform k-means clustering (with $\textup{k} \in \{80, 160\}$) on user features and build the so-called similarity matrix $\bW$. 
We use the dot-product of features between two clusters to populate $\bW$. We then introduce sparsity as needed and normalize $\bW$ column-wise.

As described earlier in the introduction, the Yahoo! Today Module (2009) dataset is considerably large 
with over $45$ million data points, $20$ candidate articles, 
and a total of $270$ articles over the course of a ten-day period. The payoff of the interaction of the user with the recommended article is 0-1, which is called a {\it click}. Hence, in this set of experiments, we use click-through-rate (CTR) to evaluate the performance of all algorithms. We normalize the performance of different algorithms by a random algorithm (which recommends each article with the same probability in each period).

The article that will be shown to a user is selected from the set of available articles at their time of arrival (i.e., we show only one out of 20). Although we observe the features of the user, we do not directly use them. We assign each user to the nearest cluster and use the features of that cluster's centroid. Then, 
if and only if the recommended article matches the historical data, the corresponding reward is collected and the model is updated. Otherwise, we do not perform any updates. 

\begin{figure}[h]
    \centering
    \subfloat[Cluster size $160$, Sparsity $100\%$]{
    \includegraphics[width=0.47\linewidth]{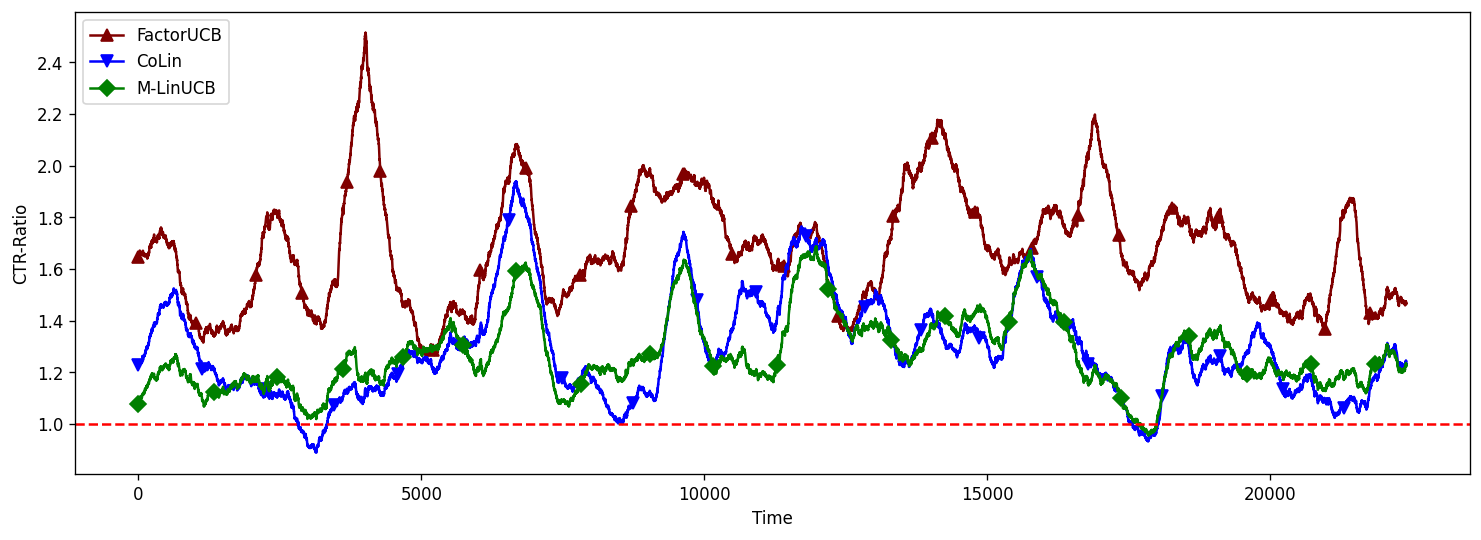}
    }\quad
    \subfloat[Cluster size $80$, Sparsity $100\%$]{
    \includegraphics[width=0.47\linewidth]{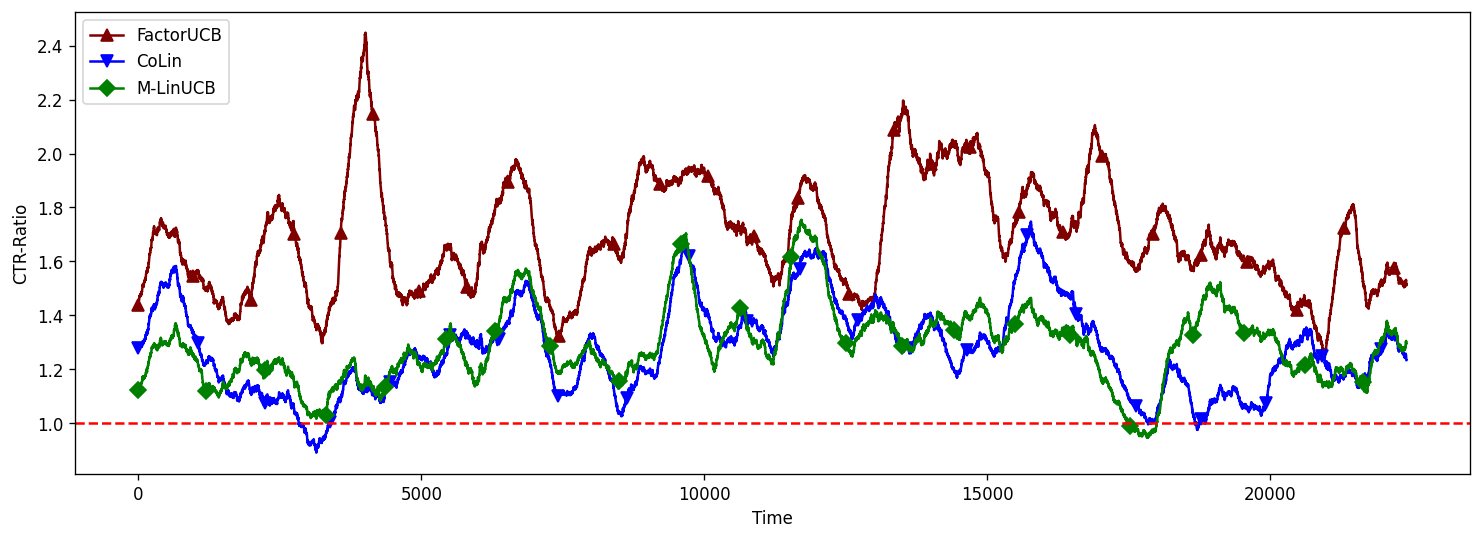}
    }\\
    \subfloat[Cluster size $160$, Sparsity $50\%$]{
    \includegraphics[width=0.47\linewidth]{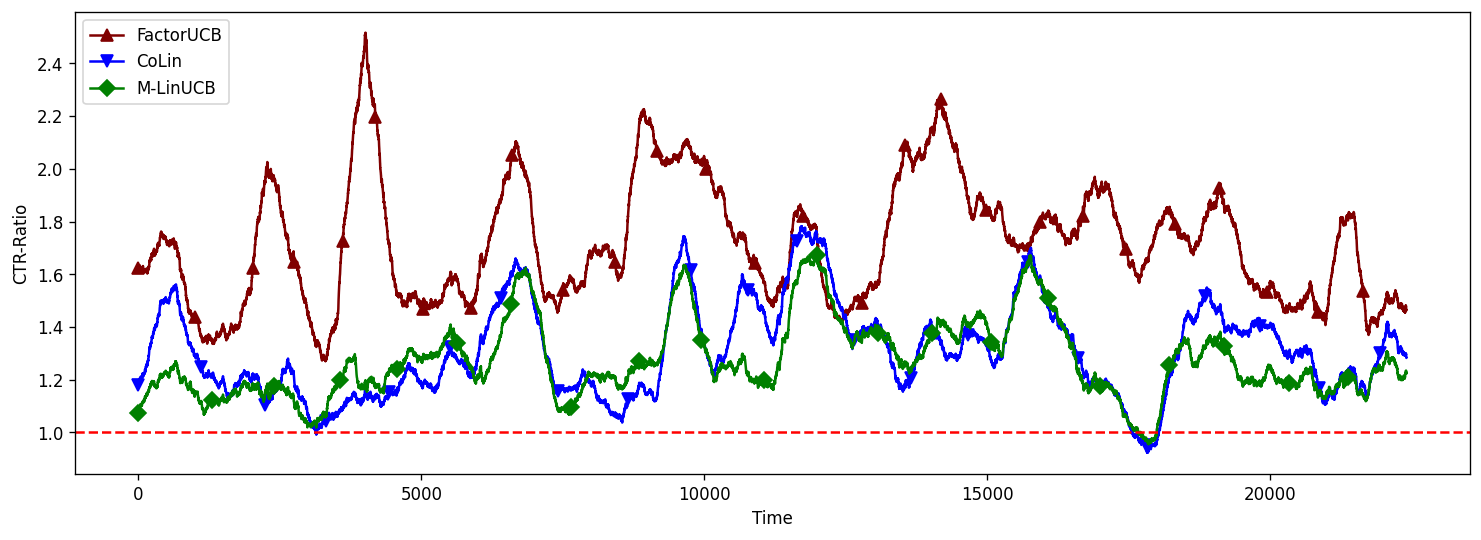}
    }\quad
    \subfloat[Cluster size $80$, Sparsity $50\%$]{
    \includegraphics[width=0.47\linewidth]{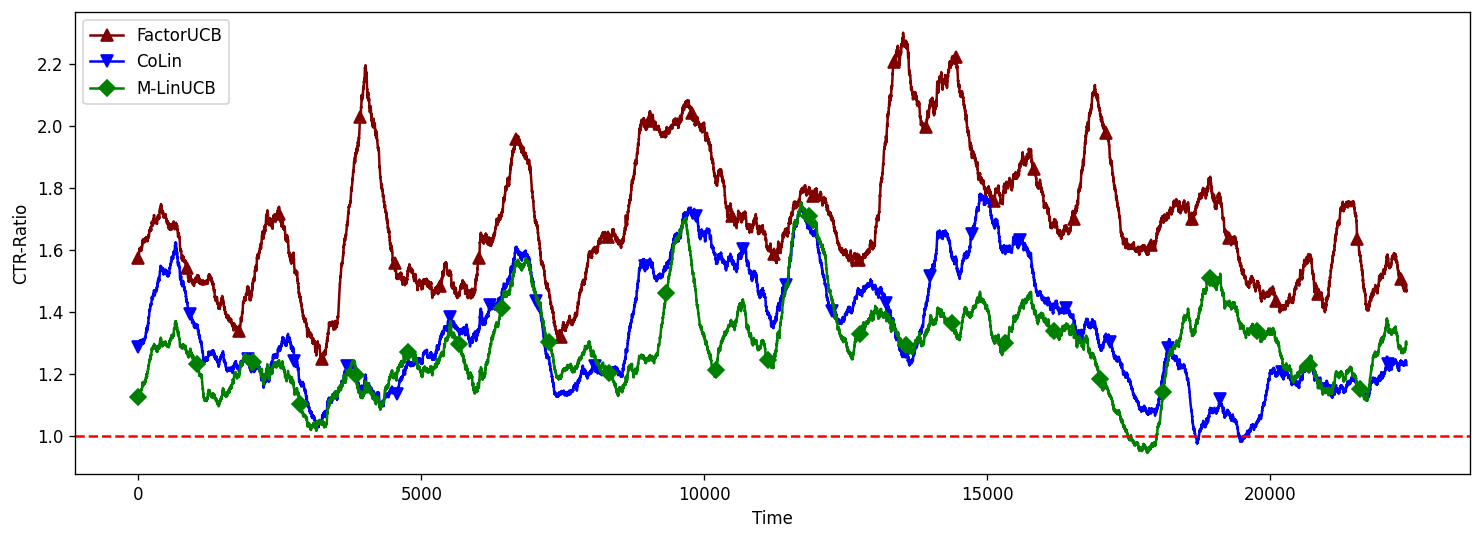}
    }\\
    \subfloat[Cluster size $160$, Sparsity $25\%$]{
    \includegraphics[width=0.47\linewidth]{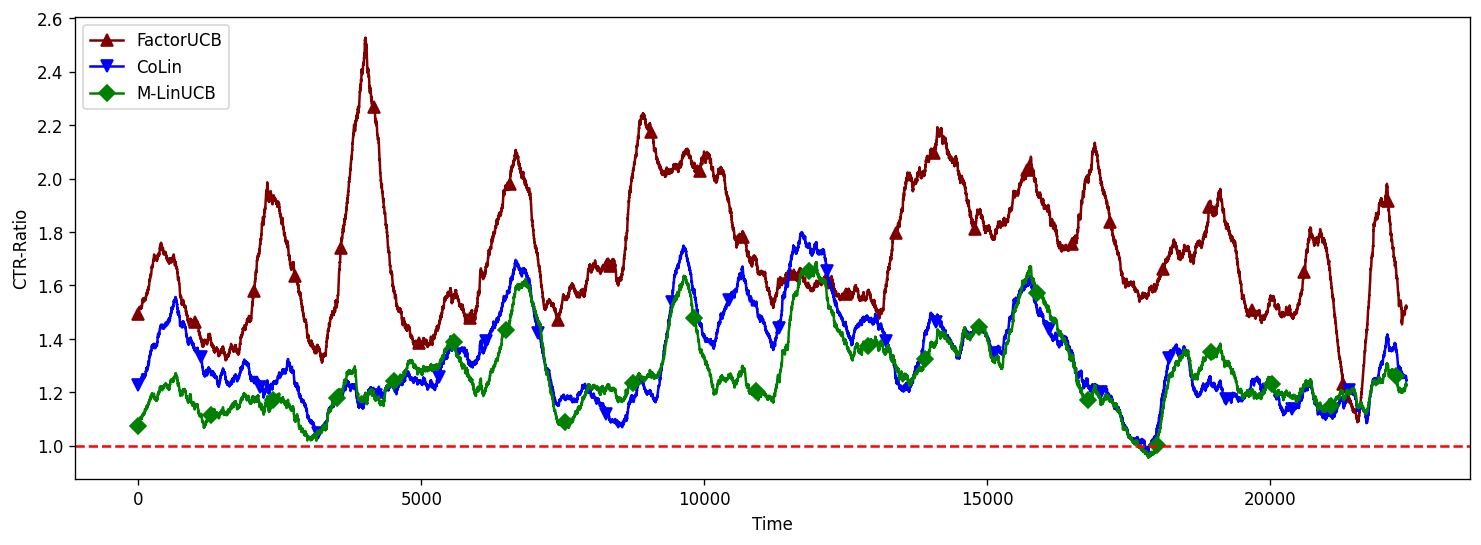}
    }\quad
    \subfloat[Cluster size $80$, Sparsity $25\%$]{
    \includegraphics[width=0.47\linewidth]{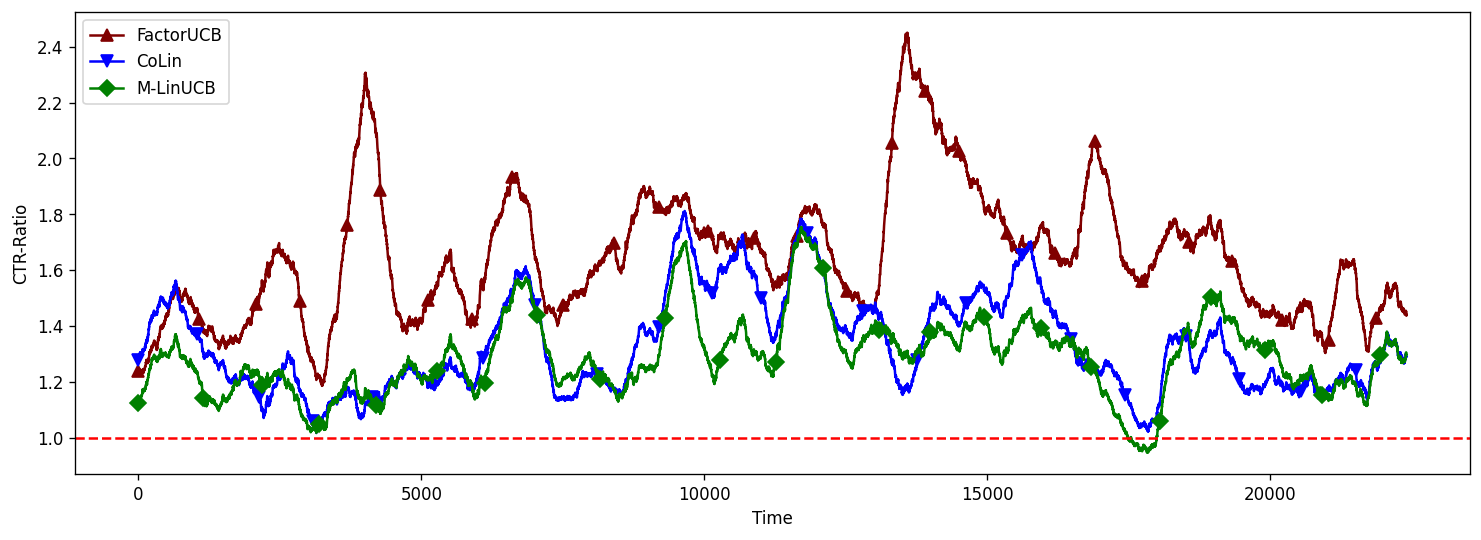}
    }
    \caption{FactorUCB, CoLin, and $\textup{M}$-LinUCB. $\alpha=0.5$. $\alpha_1=\alpha_2=0.375$ for all FactorUCB.\label{fig:yahoo_main_alphaHalf}}
\end{figure}

\begin{figure}[h]
    \centering
    \subfloat[Cluster size $160$, Sparsity $100\%$]{
    \includegraphics[width=0.47\linewidth]{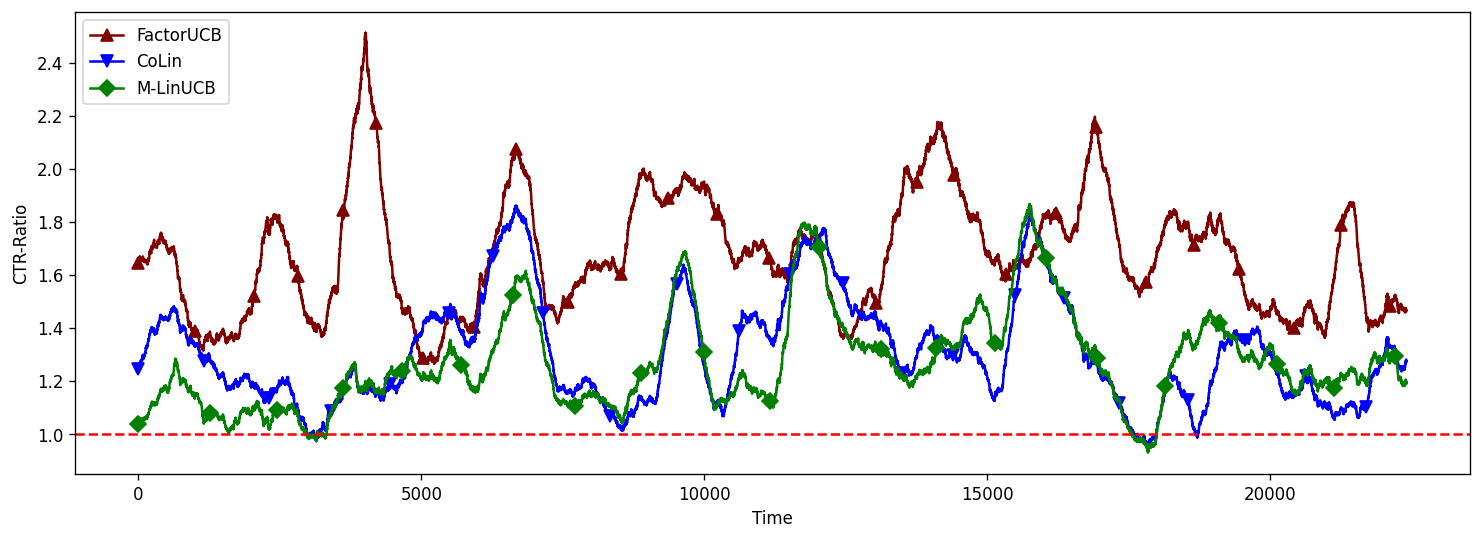}
    }\quad
    \subfloat[Cluster size $80$, Sparsity $100\%$]{
    \includegraphics[width=0.47\linewidth]{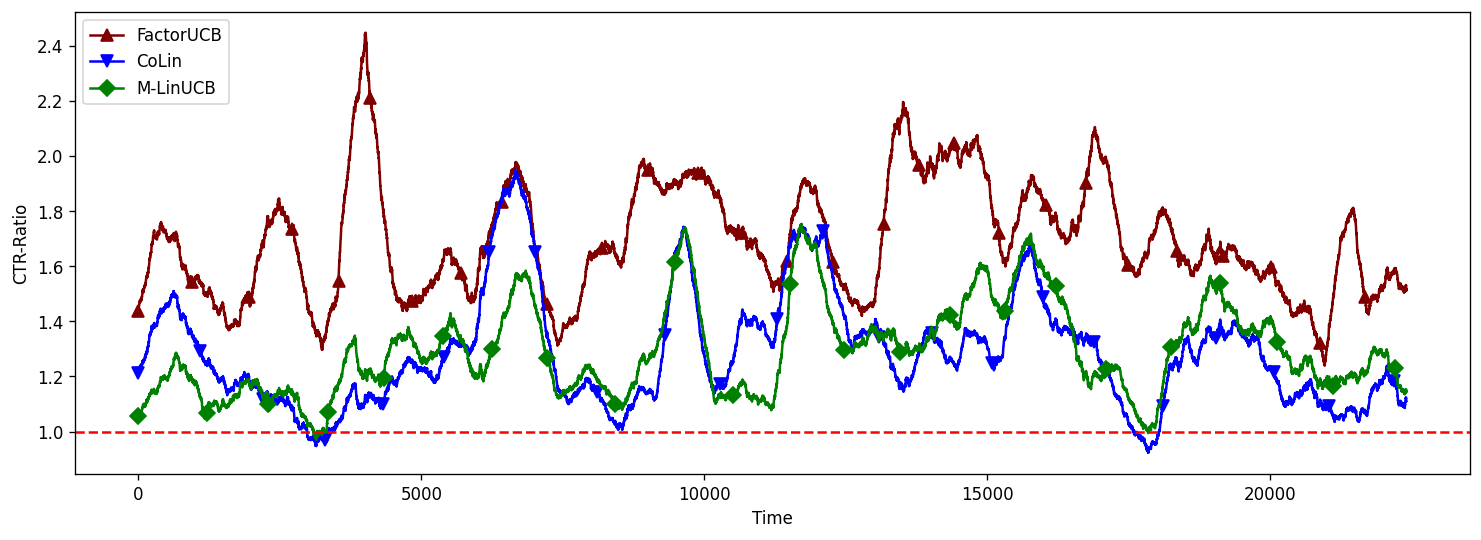}
    }\\
    \subfloat[Cluster size $160$, Sparsity $50\%$]{
    \includegraphics[width=0.47\linewidth]{figs/Yahoo_160UserClusters_Sparsity50_alpha0.5.png}
    }\quad
    \subfloat[Cluster size $80$, Sparsity $50\%$]{
    \includegraphics[width=0.47\linewidth]{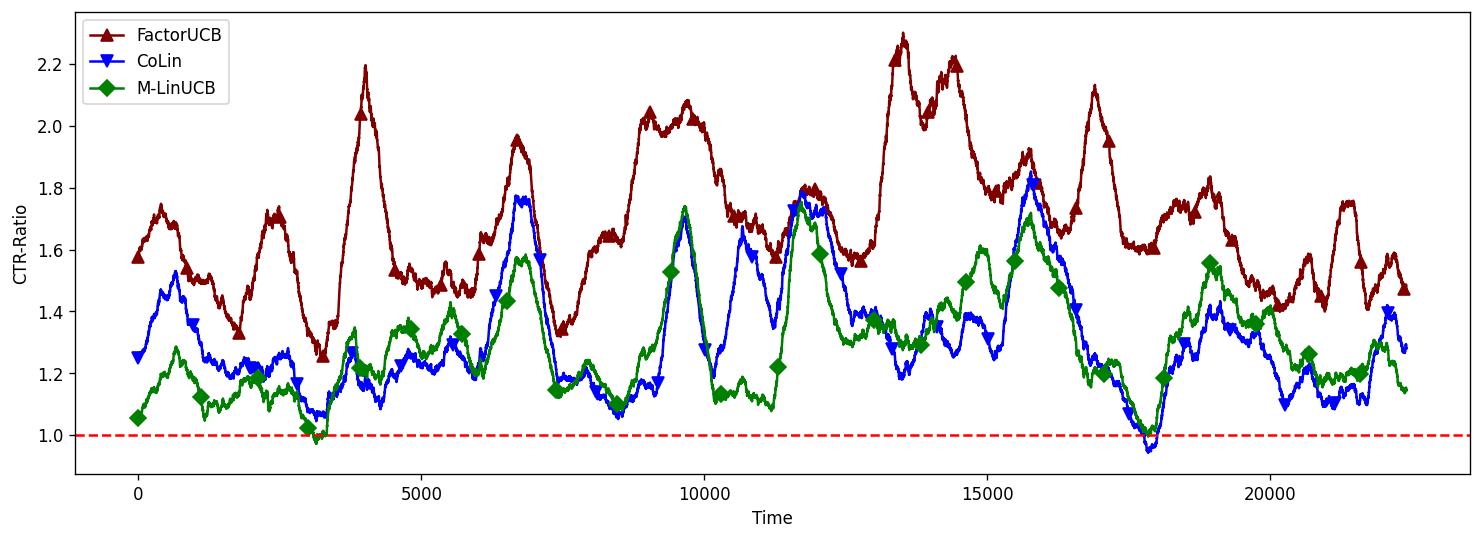}
    }\\
    \subfloat[Cluster size $160$, Sparsity $25\%$]{
    \includegraphics[width=0.47\linewidth]{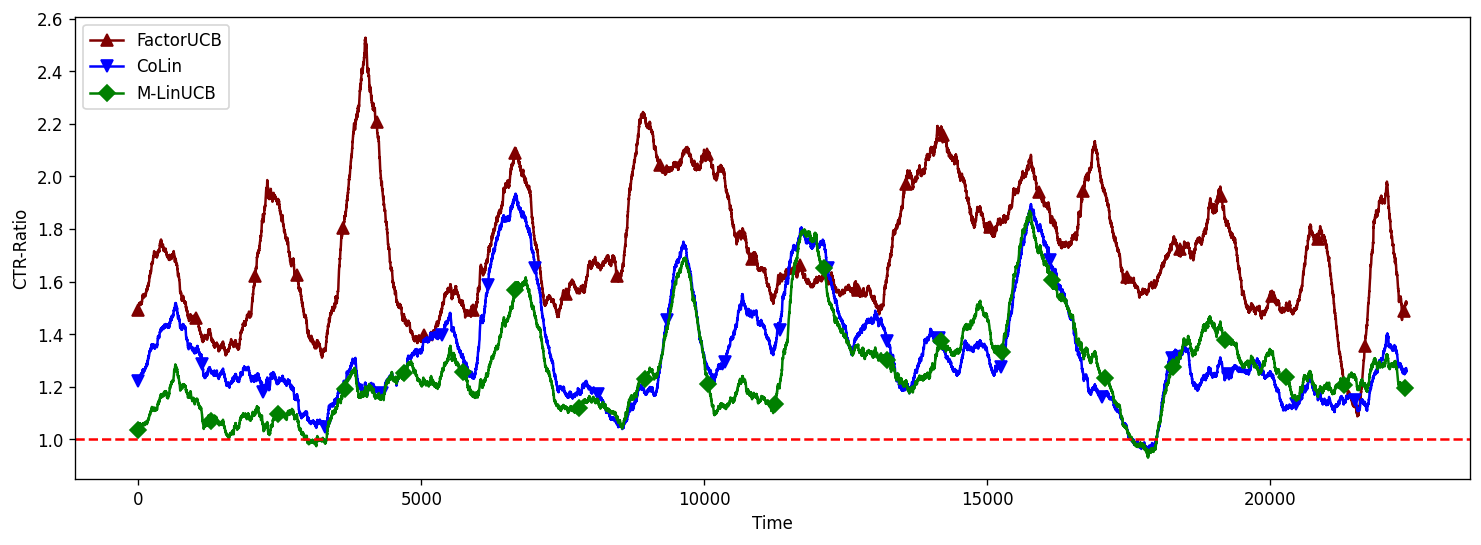}
    }\quad
    \subfloat[Cluster size $80$, Sparsity $25\%$]{
    \includegraphics[width=0.47\linewidth]{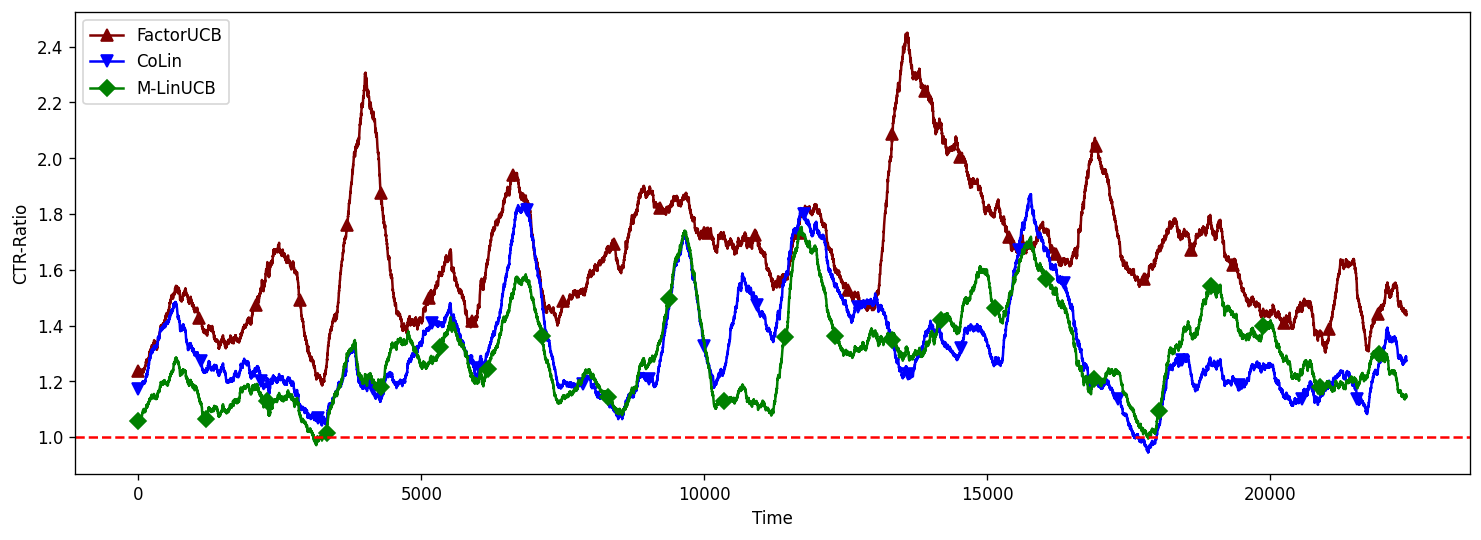}
    }
    \caption{FactorUCB, CoLin, and M-LinUCB. $\alpha=1$. $\alpha_1=\alpha_2=0.375$ for all FactorUCB.\label{fig:yahoo_main_alpha1}}
\end{figure}

\begin{figure}[h]
    \centering
    \subfloat[FactorUCB and CoLin]{
    \includegraphics[width=0.48\linewidth]{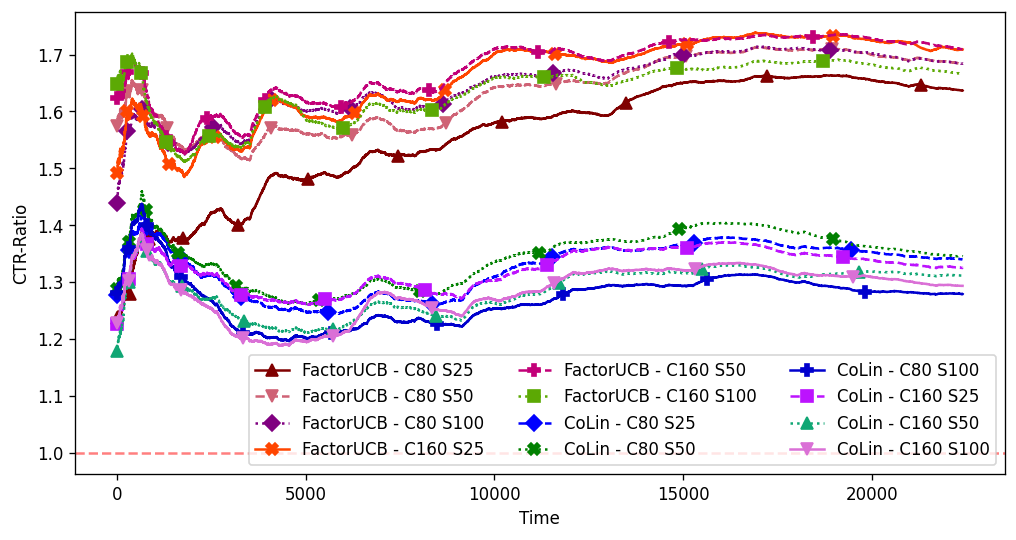}
    }\quad
    \subfloat[CoLin and M-LinUCB]{
    \includegraphics[width=0.48\linewidth]{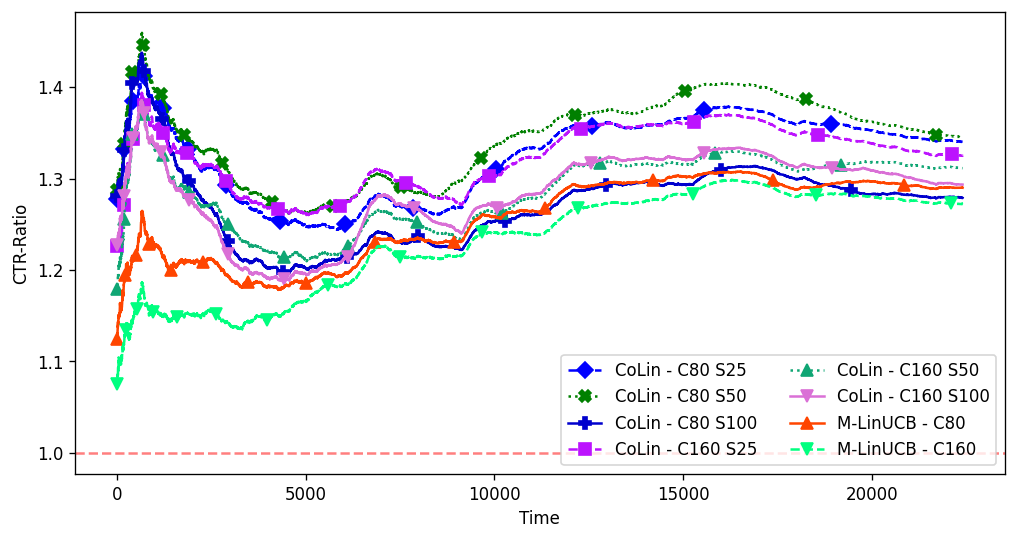}
    }
    \caption{Cumulative CTR performance for FactorUCB, CoLin, and M-LinUCB; $\alpha=0.5$, $\alpha_1=\alpha_2=0.375$.\label{fig:yahoo_overall_avg_aHalf}}
\end{figure}

\begin{figure}[h]
    \centering
    \subfloat[FactorUCB and CoLin]{
    \includegraphics[width=0.48\linewidth]{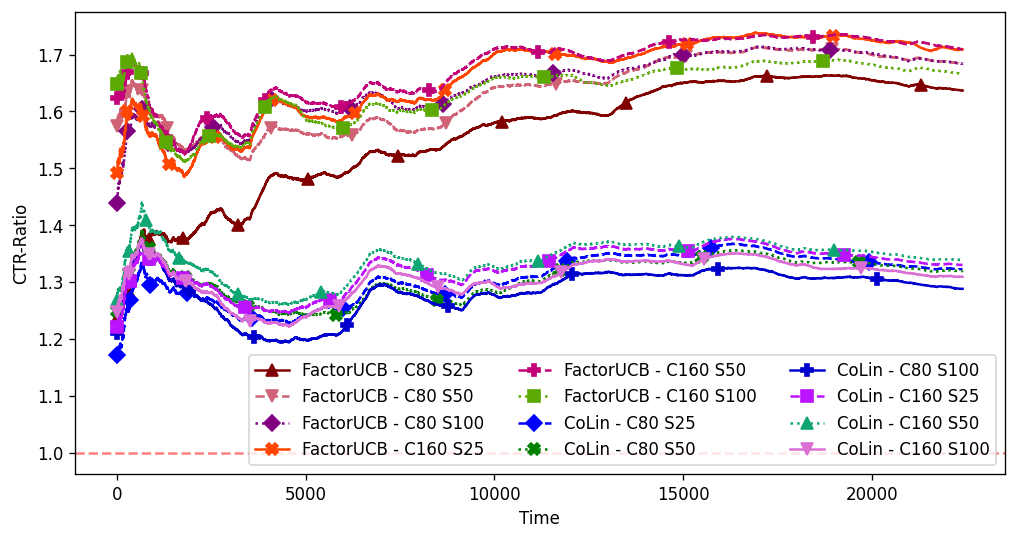}
    }\quad
    \subfloat[CoLin and M-LinUCB]{
    \includegraphics[width=0.48\linewidth]{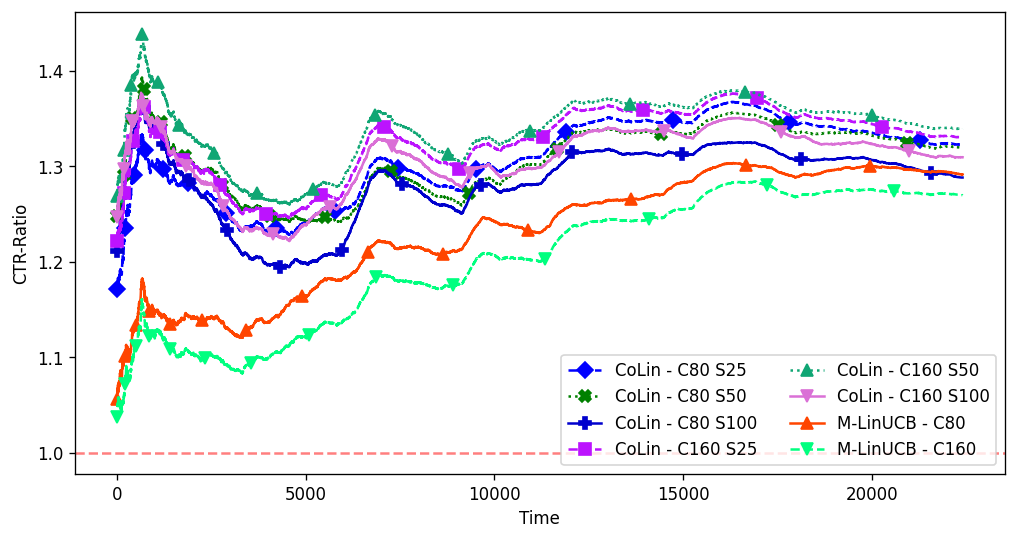}
    }
    \caption{Cumulative CTR performance for FactorUCB, CoLin, and M-LinUCB; $\alpha=1$, $\alpha_1=\alpha_2=0.375$.\label{fig:yahoo_overall_avg_a1}}
\end{figure}

In Figures \ref{fig:yahoo_main_alphaHalf} and \ref{fig:yahoo_main_alpha1}, we report rolling averages of CTR values. We calculated the average CTR every 2,000 time periods. This process generated 22,905 data points, which we used to create the figures. Using a rolling average of $500$ to average the CTR values, we generate Figures \ref{fig:yahoo_main_alphaHalf} and \ref{fig:yahoo_main_alpha1}. (Time, the $x$-axis, goes from zero to $22,405$.) In Figures \ref{fig:yahoo_overall_avg_aHalf}, \ref{fig:yahoo_overall_avg_a1}, and \ref{fig:yahoo_overall_avg_perAlg}, we report the cumulative average CTR values.

\subsubsection{Comparing Experiment Results by Varying the Cluster Size and the Sparsity}\label{subsubsec:clusterSize}
We first consider the effect of changing the cluster sizes, as this affects all algorithms. Changing the sparsity does not affect M-LinUCB, and changing the $\alpha$ value does not affect FactorUCB. Note that FactorUCB uses $\alpha_1$ and $\alpha_2$ and we set $\alpha_1=\alpha_2=0.375$ in these experiments. We present the experiments where we vary $\alpha_1$ and $\alpha_2$ values for FactorUCB in Section \ref{sec:varyAlpha_FactorUCB}.

Figures \ref{fig:yahoo_main_alphaHalf} and \ref{fig:yahoo_main_alpha1} are organized by $\alpha$: Figure \ref{fig:yahoo_main_alphaHalf} has $\alpha=0.5$ and Figure \ref{fig:yahoo_main_alpha1} has $\alpha=1$. Recall that we normalize the performance of all algorithms by a random policy, where the random algorithm recommends each candidate article with the same probability in each period.
In both figures, the figures on the left has cluster size $160$, and the figures on the right has cluster size $80$. 
We denote a cluster size $160$ as C$(160)$ and a sparsity of $100\%$ as S$(100)$, to refrain from continuously repeating those words. (We leave them as they are in the figure captions.)

We observe that FactorUCB performs best for C$(160)$ and S$(50)$, with C$(160)$ and S$(25)$ being a close second: See Figure \ref{fig:yahoo_overall_avg_perAlg}(a). FactorUCB with C$(160)$ and S$(50)$ outperforms the random policy by $67.5\%$, and the best CoLin by $24.6\%$ (see Figure \ref{fig:yahoo_overall_avg_aHalf}(a)). The best CoLin outperforms the best M-LinUCB by $7\%$. From this, we can immediately reason that the marginal value of the similarity matrix $\bW$ is low since the improvement of CoLin over M-LinUCB is fairly small compared to improvement of FactorUCB over M-LinUCB. Especially, CoLin with C$(80)$ and S$(100)$ frequently under-performs $\textup{M}$-LinUCB in Figure \ref{fig:yahoo_main_alpha1}(b) (and its cumulative CTR average eventually dips below that of M-LinUCB, see Figure \ref{fig:yahoo_overall_avg_aHalf}(b)).

CoLin improves for C$(80)$ when the sparsity is reduced, or the cluster size is increased with any sparsity level. This suggests that a cluster size of $80$ with full sparsity is of low quality. Hence, we do not test a cluster size smaller than $80$ in our experiments. The original FactorUCB work \citep{wang2017factorization} also notes that, for the same Yahoo! Today Module dataset, FactorUCB with a trivial similarity matrix $\bW$ (i.e., $\bW = \mathbf{I}$) outperforms FactorUCB with a non-trivial $\bW$. While we do not test that particular configuration, comparing CoLin to $\textup{M}$-LinUCB helps in arriving to a similar observation. Therefore, introducing a static $\bW$ only may not be worth for the performance improvement as it brings with a high computation complexity. We present another empirical proof for the potential misspecification of $\bW$ in the next subsection.

FactorUCB performs best for C$(160)$ when S$(50)$, and for C$(80)$ it performs the best when S$(100)$. This suggests that, while a larger cluster size with full connectivity may degrade the performance, a smaller cluster size with any sparsity level is not a match for a larger cluster size. (See Figure \ref{fig:yahoo_overall_avg_perAlg}(a) for a clear picture.) Therefore, we argue that introducing sparsity and allowing a loosely connected forest (or clique) structure with large cluster sizes generate the best improvement, bringing about an improvement of $2.5\%$ over its smaller cluster size counterpart. This aligns with our intuition: while a larger cluster size allows for more flexible modeling, full connectivity might spread irrelevant information to some users. This could hurt the performance for users who would benefit from more targeted information sharing.\footnote{A larger cluster size with smaller sparsity, heuristically, only preserves the strong-enough relationships.} Coincidentally, we observe that the performance improves by $2.3\%$ when sparsity is halved for C$(160)$ from S$(100)$ to S$(50)$.


Lastly, partly in contrast to intuition above, we observe that M-LinUCB performs better with a smaller cluster size, independent of the value of $\alpha$. We argue that this is due to the information sparsity and the absence of collaboration in M-LinUCB. Increasing the cluster size means that each cluster receives less number of reward signals. Since the click data is sparse, doubling the cluster size means even sparser data for each cluster. We discuss this more in Section \ref{sec:discussion} and provide more context.

\subsubsection{Comparing Experiment Results by Varying $\alpha$}
Next, we examine the effect of varying $\alpha$. This parameter dictates the intensity of exploration. We do not mention FactorUCB in this subsection, as its performance is independent of $\alpha$ in the main sets of experiments. We present another set of experiments where we vary $\alpha$ for FactorUCB in Section \ref{sec:varyAlpha_FactorUCB}. 

\begin{figure}[h]
    \centering
    \subfloat[FactorUCB]{
    \includegraphics[width=0.48\linewidth]{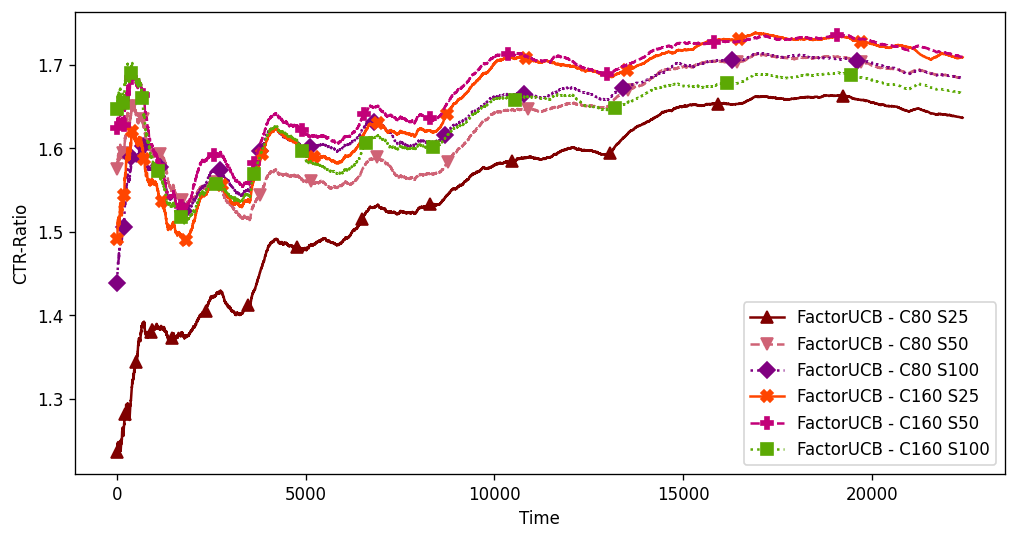}
    }\quad
    \subfloat[M-LinUCB]{
    \includegraphics[width=0.48\linewidth]{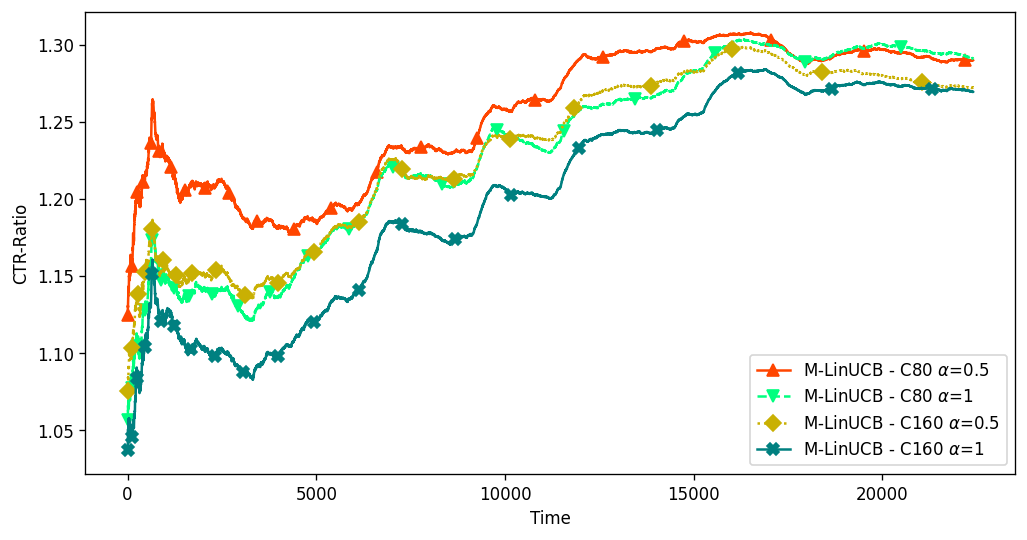}
    }\quad
    \subfloat[CoLin, $\alpha=0.5$]{
    \includegraphics[width=0.48\linewidth]{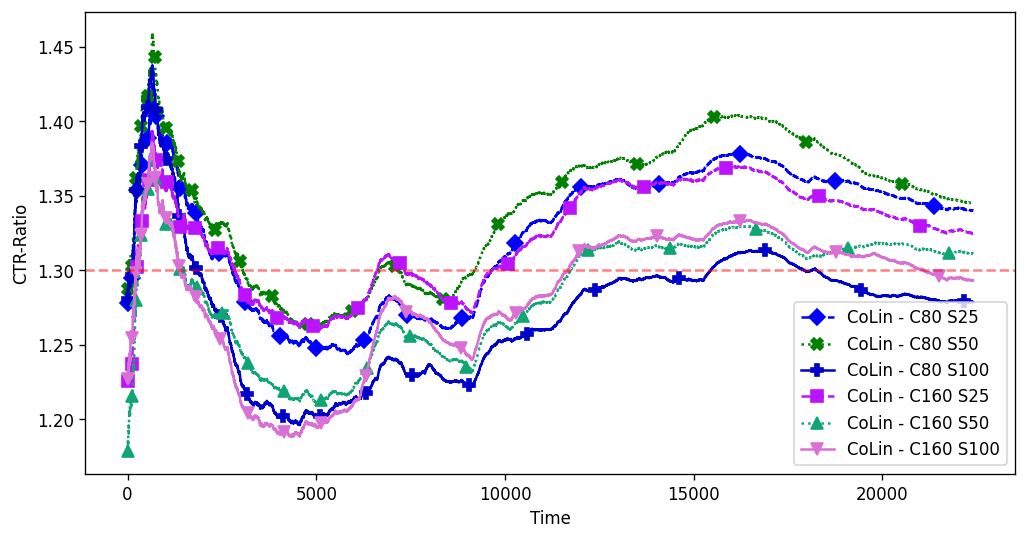}
    }\quad
    \subfloat[CoLin, $\alpha=1$]{
    \includegraphics[width=0.48\linewidth]{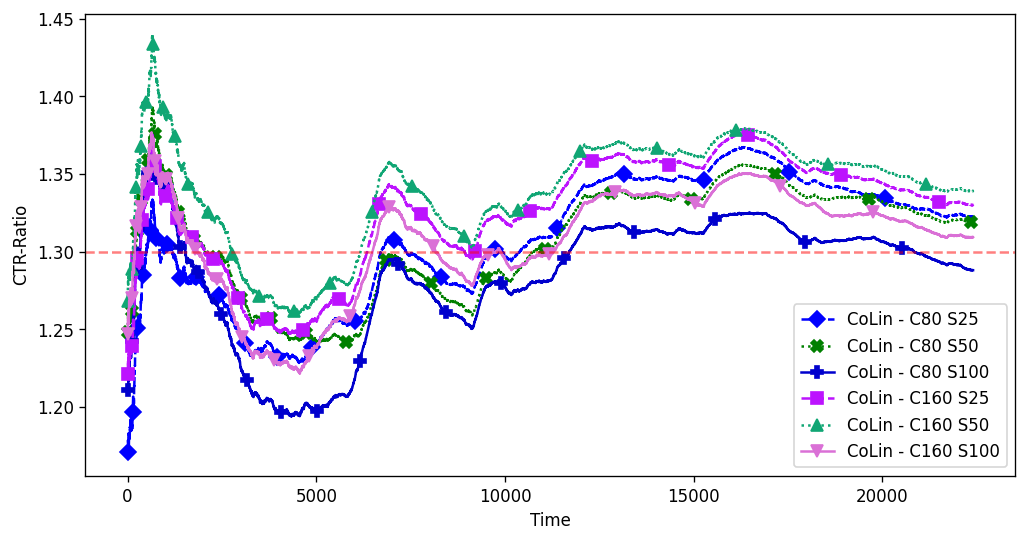}
    }
    \caption{Cumulative CTR performance with all parameter settings for each algorithm.\label{fig:yahoo_overall_avg_perAlg}}
\end{figure}

In Figure \ref{fig:yahoo_main_alphaHalf}, CoLin performs best for C$(80)$ and S$(50)$: It outperforms the random policy by $34.4\%$. When $\alpha$ is increased from $0.5$ to $1$ in Figure \ref{fig:yahoo_main_alpha1}, CoLin performs best for C$(160)$ and S$(50)$: It outperforms the random policy by $33.9\%$. This difference can be seen clearly in Figure \ref{fig:yahoo_overall_avg_perAlg}(d). On average, CoLin outperforms the random policy by $30.6\%$ and $31.2\%$ when $\alpha=0.5$ and $\alpha=1$, respectively. That is, while the best performance is when $\alpha=0.5$, so is the worst performance. When $\alpha=1$, the performance is more stable across different cluster sizes and sparsity values. This may suggest that the similarity matrix $\bW$ is slightly misspecified and tendency to explore more is acting as a correction, reducing the variance due to hyperparameter selection. 

Regarding the above point, potential misspecification of the similarity matrix $\bW$, becomes more clear when we consider M-LinUCB and how its performance changes due to $\alpha$. Note that M-LinUCB does not utilize $\bW$, and therefore is independent of the choice of sparsity parameter. 

On average, M-LinUCB outperforms the random policy by $24.4\%$ and $21.7\%$ when $\alpha=0.5$ and $\alpha=1$, respectively. In Figure \ref{fig:yahoo_overall_avg_perAlg}(b), it can be seen that, in terms of cumulative CTR averages, the effect of varying $\alpha$ values seem to eventually disappear and the cluster size makes the most difference.

The spread of cumulative CTR performances in Figure \ref{fig:yahoo_overall_avg_perAlg}(c) is wider than those in \ref{fig:yahoo_overall_avg_perAlg}(d). As mentioned above, CoLin performs its best and worse for $\alpha=0.5$. 
In UCB-based policies, the confidence bounds gets tighter with time, and hence the frequency of taking exploratory actions decreases. If we set a small $\alpha$, then these confidence bounds gets tighter faster.
The fact that the spread is wider in Figure \ref{fig:yahoo_overall_avg_perAlg}(c) implies the {\it empirically best items} identified by CoLin are not the true best items, and CoLin is more accurate in identifying the true best items when it is allowed more room to explore (i.e., $\alpha=1$). 
Therefore, performance deterioration due to the similarity matrix $\bW$ remaining static throughout the time horizon can be remedied by increasing the exploration intensity, i.e., the value of $\alpha$ parameter.


\subsubsection{Varying the Exploration Intensity in FactorUCB}\label{sec:varyAlpha_FactorUCB}
We close the discussion on numerical results in this section by presenting the results of additional tests conducted on the selection of $\alpha_1$ and $\alpha_2$ parameters for FactorUCB. 
We run two additional tests where we compare the original setting of $\alpha_1$ = $\alpha_2$ = $0.375$ with the following two settings: (1) $\alpha_1 = 2\alpha_2 = 0.5$, and, (2) $2\alpha_1 = \alpha_2 = 0.5$. 
We set cluster size to $160$ and sparsity to $50\%$.

\begin{figure}[h]
    \centering
    \subfloat[Rolling performance]{
    \includegraphics[width=0.47\linewidth]{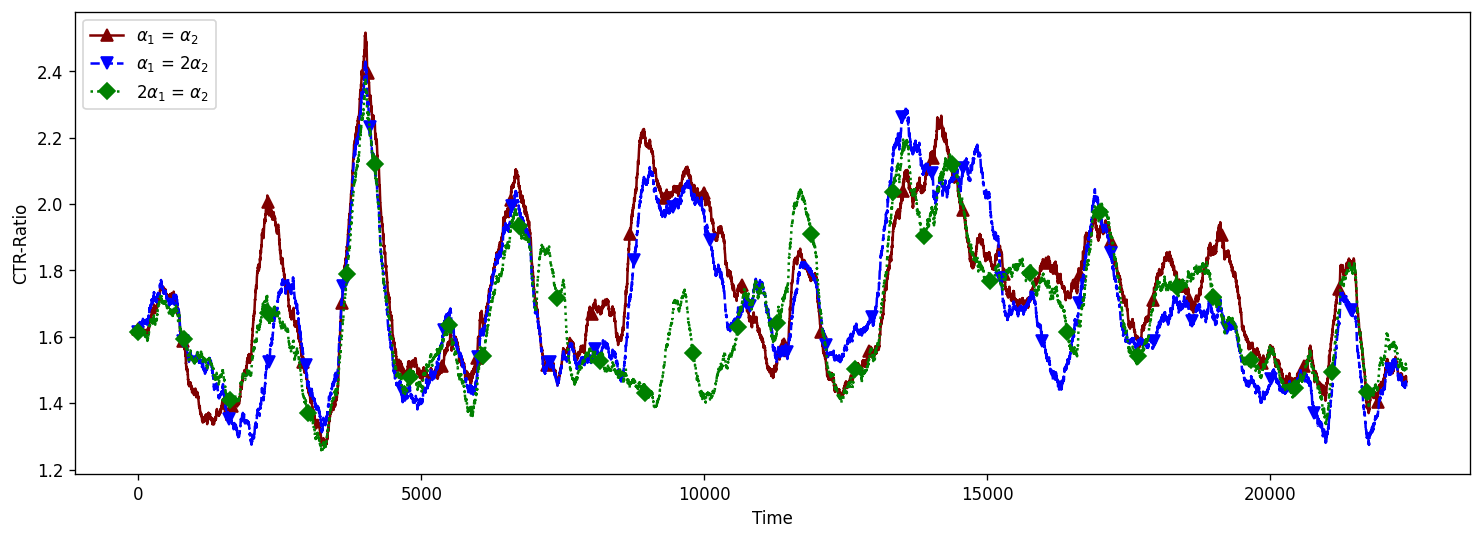}
    }\quad
    \subfloat[Cumulative CTR performance]{
    \includegraphics[width=0.47\linewidth]{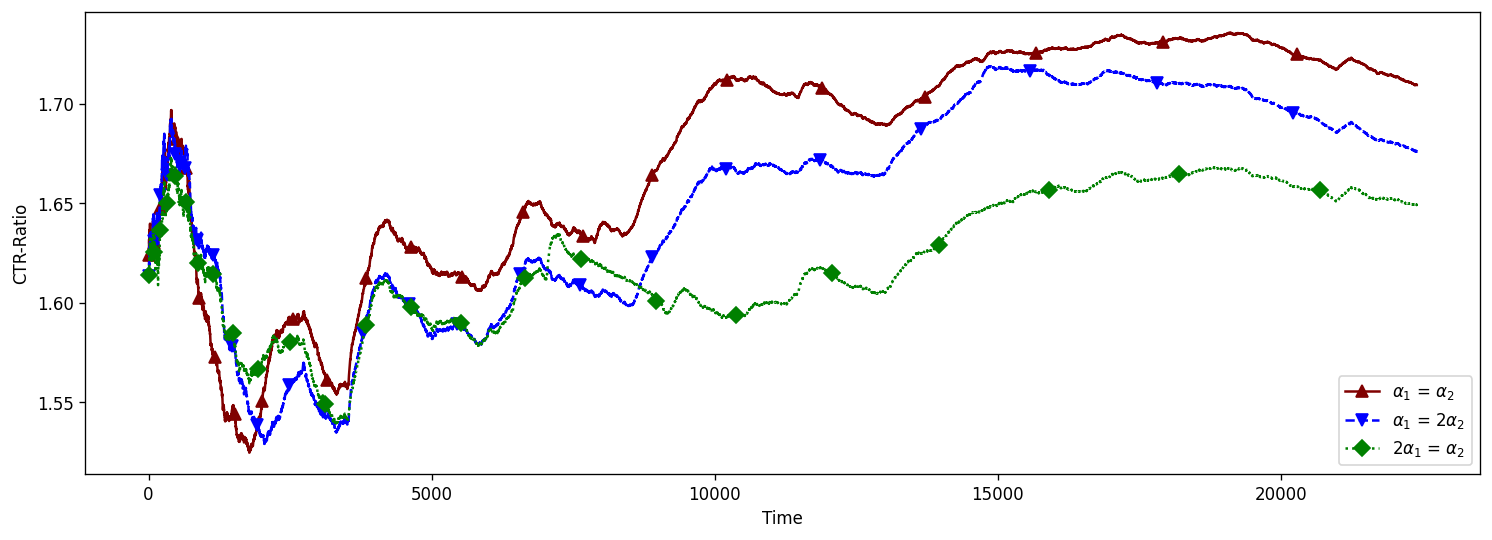}
    }
    \caption{Performance comparison for varying values of $\alpha_1$ and $\alpha_2$ for FactorUCB.\label{fig:factorUCB_varyAlpha}}
\end{figure}

In Figure \ref{fig:factorUCB_varyAlpha}(a), we have the rolling average performance of CTR, in similar fashion to Figures \ref{fig:yahoo_main_alphaHalf} and \ref{fig:yahoo_main_alpha1}. In Figure \ref{fig:factorUCB_varyAlpha}(b), we have the cumulative performance, in similar fashion to Figures \ref{fig:yahoo_overall_avg_aHalf}, \ref{fig:yahoo_overall_avg_a1}, and \ref{fig:yahoo_overall_avg_perAlg}.

The original instance at $\alpha_1$ = $\alpha_2$ = $0.375$ with cluster size $160$ and sparsity $50\%$ performs the best among the earlier FactorUCB results, outperforming the random policy by $67.5\%$. 
The instance with $\alpha_1 = 2\alpha_2 = 0.5$ outperforms the random policy by $65.2\%$, and the instance with $2\alpha_1 = \alpha_2 = 0.5$ outperforms the random policy by $62.2\%$.

In Figure \ref{fig:factorUCB_varyAlpha}(a), we can see that the $2\alpha_1 = \alpha_2 = 0.5$ instance lags behind most of the time while the $\alpha_1 = 2\alpha_2 = 0.5$ instance matches the original instance's behavior more closely. Although, in Figure \ref{fig:factorUCB_varyAlpha}(b), we observe that the $\alpha_1 = 2\alpha_2 = 0.5$ instance, while getting very close to the original instance around time $15,000$, eventually diverges. 
These results confirm our setting of $\alpha_1$ = $\alpha_2$ = $0.375$ for FactorUCB. For the Yahoo dataset, we achieve the best performance out of FactorUCB using equal $\alpha$ values.

Nevertheless, these tests open up new avenues for future research, such as (1) quantifying the potential gains that can be obtained by dynamically varying the $\alpha$ values in real time, 
and, (2) varying $\alpha$ values and using them as a lever to control for the misspecification of similarity matrix $\bW$. See \cite{tor_ADAUCB} for an approach similar to the former point above in the context-free multi-armed bandit setting.



\subsubsection{Discussion on Collaboration and Algorithms}\label{sec:discussion} 

Lastly, we close this section with additional discussion on the use of collaboration in previously described algorithms. 
Recall that the basic model we use is a function defined on the user (bandit) parameters and item features, returning a real-valued score:
\begin{align}
    g: \mathbb{R}^{d_u} \times \mathbb{R}^{d_a} \mapsto \mathbb{R}; g(\btheta, \bx) \mapsto p,
\end{align}
where the actions taken by the policy depends on the score $p$, $\bx$ is given and the (bandit) parameter $\btheta$ is to be learned. Each $\btheta$ may be different for distinct users.

In offline (batch) learning, the class imbalance problem (i.e., the problem of certain class(es) being represented considerably less than the other(s)) usually brings sparsity problems, hindering the learning of true parameters that prescribes the best model within a specific function class. 
The same imbalance issue becomes harder to solve in online settings because, additionally, the data that is collected at time $t_2 > t_1$ does not inform the model at time $t_1$. Therefore, if there is no collaboration among users, having access to other users' functions is not helpful in a model setting where a separate function is defined for each user and their interactions with items.

We numerically observe that a collaboration structure helps remedy this temporal disconnection while learning about users with different characteristics, i.e., one user's experience may inform another. The similarities between the preferences of distinct users can be modeled and used to accelerate learning in online settings. 
Therefore, we argue that a collaboration structure brings two main advantages. The first is speeding up learning even when the similarity matrix $\bW$ may be misspecified. See Figures \ref{fig:yahoo_overall_avg_aHalf}(b) and \ref{fig:yahoo_overall_avg_a1}(b) to compare early stages of CoLin to early stages of M-LinUCB, even though they eventually converge to each other. 
The second advantage is making exploration more targeted by bootstrapping, i.e., updating (the estimate of) the bandit parameter $\btheta_1$ using (the estimate of) the bandit parameter $\btheta_2$, given that the model dictates that users $1$ and $2$ share similarities. 
Both of these advantages also align the policies that utilize a collaboration structure the reinforcement learning literature (or dynamic programming): ``Update estimates based in part on other learned estimates, without waiting for a final outcome,'' \citep{RLbook}. 

Note that, if starting from scratch with no access to historical data, the similarity matrix $\bW$ cannot be set at the beginning of the planning horizon because there is no information on users at time $t=0$. 
Therefore, a core assumption of this modeling approach is having access to historical data collected in a setting similar to the modeling approach. 
Nonetheless, additional modeling considerations, such as the use of latent factors in FactorUCB, may help in recovering the information lost, or excluded, while building $\bW$ at time $t=0$.

Lastly, recall that the Yahoo! Today Module dataset is made up of user interactions with articles over the course of a ten-day period. Across and within these days, the popularity of articles changes. Therefore, in Figures \ref{fig:yahoo_main_alphaHalf} and \ref{fig:yahoo_main_alpha1}, we observe peaks and valleys for each algorithm. We usually run into peaks when a popularity shift does not occur for a long enough time so that the models can catch up, and run into valleys soon after such a shift occurs. 
Here, it is important to point out that FactorUCB is less robust to these popularity shifts than CoLin and $\textup{M}$-LinUCB. Usually, FactorUCB quickly peaks but also loses momentum fast, and its performance occasionally gets close to those of CoLin and $\textup{M}$-LinUCB. 

We argue that high variance of FactorUCB is potentially due to two structural elements of the algorithm: (1) latent factors are {\it learned} through interactions with historical data, and hence, are high variance; and (2) even though the latent factors help with propagating existing information, they are not as helpful for brand new articles with insufficient information, which are also represented by their own latent factors. 
For the first point, one can control the consecutive changes made in latent parameter updates. And for the second point, one can {\it warm-start}\footnote{By warm-starting, we mean non-trivial initialization of unknown parameters using historical data.} the articles by designing a distinct similarity matrix for them.
We discuss these in more detail in Section \ref{sec:concl}.





\subsection{Time and Space Complexity}

We review 
the elements of the algorithms we analyzed that affect their performance and runtime complexity.

\paragraph{Reducing dimensionality.}
As the extent of data collected on users is abundant, many practical algorithms focus on reducing number of features used in modeling, and, ideally, only focus on the most efficient subset of features which describe the reality the best. 
In the case of Yahoo! Today Module dataset, this is done by the authors. The original data has $82$ features for the articles, and $1192$ features for the users. We refer readers to \cite{yahooDataset} and \cite{linucb2010} for details on dimensionality reduction.

Naturally, keeping the dataset as is and using it directly is not efficient, both computation- and memory-wise. Regardless, the publicly available version of the dataset does not contain all these features, and additionally removes user identifications. Hence, in all our experiments, we have resorted to clustering users with k-means.

\paragraph{Matrix inversions.}
Notice that the parameter covariance matrix $A$ is a square matrix with the number of rows equal to the multiplication of the  dimension of features and the number of clusters. Thus, for CoLin (Algorithm \ref{alg:colin}) and Yahoo! Today Module dataset, this matrix has close to $1,000$ rows. We need access to the inverse of this matrix at each time period, and because matrix inversion is (close to) cubic in the number of rows, we seek to avoid inverting it\footnote{And also avoid using a batch learning approach where we do not update the parameters in every period.} 
by using the fact that all of the relevant updates that include matrix inverses are rank-one, i.e., the additive update is the outer product of two vectors. (See line 15 in Algorithm \ref{alg:m-linucb}, line 14 in Algorithm \ref{alg:colin}, lines 14 and 17 in Algorithm \ref{alg:factorucb}.) 
Also note that none of these algorithms need access to the original (non-inverted) versions of the matrices for calculations, only to their inverses. 
Therefore, we only keep track of the inverse of those matrices and use Sherman-Morrison formula \citep{ShermanMorrison} as follows.
\begin{remark}
Fix $n \in \mathbb{N}$ and let $A \in \mathbb{R}^{n\times n}$ be an invertible matrix, and $u,v \in \mathbb{R}^{n}$ be column vectors. Then,
\begin{align}
    \left( A + uv^{\textup{T}} \right)^{-1} = A^{-1} - \frac{1}{1 + v^{\textup{T}}A^{-1}u} \left(A^{-1}u \right)\left(v^{\textup{T}}A^{-1}\right), \label{eq:SM}
\end{align}
if and only if $1 + v^{\textup{T}}A^{-1}u\neq0$. We follow the notation in Equation~\ref{eq:SM} in performing the rank-one updates in our implementations to attain the speed efficiencies.
\end{remark}


\section{Conclusion}\label{sec:concl}

In this paper, we survey and experiment with online learning algorithms tailored for the multi-armed bandit models aimed at solving the reward signal sparsity issue that is prevalent in many recommender systems. We emphasize that collaboration is an important ingredient for harnessing the power of shared preferences across users and therefore improving data efficiency.

We examine Collaborative Filtering (CF) \citep{CF_orig} as a popular approach that informs the recommender systems in injecting collaboration by finding relationships between users or between items, or, by embedding users and items to a shared latent factor space. We regard these approaches as {\it offline}, because they generally use historical data to build production models, and we focus on multi-armed bandit algorithms that utilize a similar perspective. 
CoLin \citep{wu2016contextual} injects collaboration by finding relationships between users, and FactorUCB \citep{wang2017factorization} builds upon CoLin by introducing a latent factor space that, instead of embedding items and users through their features, explores augmenting the learning process by introducing latent factors to items. 

Both CF-based and online learning algorithms examine help in cold start cases, i.e., when a new item (user) arrives to the platform with no history of interactions and the learner only has access to its (their) features. 
Using collaboration, the learner can effectively {\it warm-start}\footnote{Recall that, by warm-starting, we mean non-trivial initialization of unknown parameters using historical data.} 
these algorithms by finding similarities between the new item (user) and the existing items (users). 
Moreover, the collaboration gets stronger as more data is collected, further enhancing the policy of the learner. Although, note that this assumes that new data is informative and conducive to learning.

We observe the following expected trend in our experiments: M-LinUCB performs the worst among these three algorithms because it does not utilize any collaboration, and therefore, is less data efficient. We especially see the effect of data sparsity and data efficiency issues when we observe performance loss in M-LinUCB as we increase the cluster size. Intuition suggests that a larger cluster size should produce a better specified model of the data, but the empirical results show that increasing the cluster size hurts. 
We argue this is due to the sparsity of the clicks in the Yahoo! Today Module dataset. As we increase the cluster size, we end up with an even sparser data per cluster. 

CoLin improves upon M-LinUCB because it enables managing data sparsity through collaboration. Nonetheless, we argue that the improvement may not worth the additional computational and space complexity because CoLin requires $10$ times the computation time of M-LinUCB. The space requirements depend on the cluster size because the parameter covariance matrix $\mathbf{A}$ is augmented by the number of clusters. 
While only doubling the computation time with respect to CoLin, FactorUCB outperforms CoLin by $25\%$, compared to CoLin's $7\%$ improvement over M-LinUCB. Thus, FactorUCB outperforms M-LinUCB by $33\%$.

We attribute the underwhelming performance of CoLin to misspecification of the similarity matrix $\bW$ and its failure to capture distribution shifts because $\bW$ is built using the user features naively and remains static throughout the time horizon. 
This misspecification may also be due to non-uniformity in user-item interactions. That is, across clusters, the user-item interaction rates may be inconsistent and imbalanced. Hence, preferring a static $\bW$ over a time-dependent one is expected to outperform the latter only when the user-item relationships are stationary across time, and all user groups are balanced in terms of interaction rates with distinct items. Even then, one can argue that a distribution shift is unavoidable. In such cases, through tracking the distribution (or population) shift in real time, updating $\bW$ with the information collected so far can be a promising alternative. This also provides a workaround for adjusting the exploration intensity through parameter tuning and gives way to data-independent implementations.

Further, CoLin or FactorUCB may be improved following the temporal design of \cite{li2019improved} and transforming the similarity matrix $\bW$ to be dynamic rather than static. We conjecture that this is more suitable for instances such as news recommendation in which the popular content quickly shifts across days, or even within hours. Although such implementations can give rise to biased algorithms, potentially creating ``echo chambers" \citep{cinelli2021echo}. 

Although a temporal design can remedy the fast-population-shift cases, we already observe FactorUCB is high variance, as discussed in Section \ref{sec:discussion}. We conjecture that this is due to two structural elements of the FactorUCB algorithm: (1) latent factors are {\it learned} through interactions with historical data, and hence, are high variance; and (2) latent factors are not as helpful for brand new articles with insufficient information. For the first point, we conjecture that controlling the change (i.e., smoothing) in latent parameter updates can help. And, for the second point, designing a distinct similarity matrix $\bW$ for the articles may be helpful as article features may be used to extract similarity values.\footnote{
Similarly, in Section \ref{sec:varyAlpha_FactorUCB}, we discuss that varying the $\alpha$ values can be a way to control the potential misspecification in $\bW$, and identifying the rules for making parameter updates to that end is an exciting future research direction.
}


Finally, we outline two more avenues for future research: 
(1) Using collaboration for federated learning \citep{federatedLinearBandit} with censored data (which may be censored in an asymmetric manner across items or users), 
and, 
(2) Bayesian approaches to tackle the delayed feedback problem \citep{delayedLinearBandit, delayedBestArm, delayedBanditGeneralizedLinear} or instances where batched decisions \citep{batchedBandit19, batchedBandit24} are required. 
For the former, distributed learning, or transfer learning with censored data, may draw attention mainly due to the ever-evolving nature of privacy protections and laws, e.g., GDPR Cookie Law.\footnote{gdpr.eu/cookies} Implementing such solutions from the lens of collaborative contextual bandits may generate more data efficient algorithms. 
And for the latter, such applications may produce more data efficient algorithms that can combine (potentially) multi-dimensional relationships for real-time implementations.

\section*{Acknowledgement}

The first author expresses gratitude for the mentorship of Arjun Ravi Kannan during his internship at Discover Financial Services, which greatly contributed to the success of this work. The views and opinions expressed in this paper are solely those of the authors and do not necessarily reflect the official policy or position of their respective employers or any affiliated organizations. Any opinions, findings, conclusions, or recommendations are those of the authors alone.


\bibliographystyle{abbrv} 
\bibliography{main}




\end{document}